%% file: main.tex
\documentclass[letter,12pt]{article}

\usepackage[ margin=1in]{geometry}
\usepackage[utf8]{inputenc}
\usepackage{graphicx}
\usepackage{caption}
\usepackage{subcaption}
\usepackage{pslatex}
\usepackage[hyphens]{url}
\usepackage[hidelinks]{hyperref}
\hypersetup{breaklinks=true}
\usepackage{apacite}
\usepackage{float}
\usepackage{multirow}
\usepackage{amsfonts,amsmath,mathtools}
\usepackage{bm}
\usepackage[dvipsnames]{xcolor}

\usepackage{booktabs,caption}
\usepackage{float}
\usepackage[flushleft]{threeparttable}

\title{\textbf{Learning the meanings of function words from grounded language using a visual question answering model}
}

\date{April 2024}
\author{Eva Portelance$^{1,2}$\thanks{Corresponding author: eva.portelance@mila.quebec} \and  Michael C. Frank$^3$ \and  Dan Jurafsky$^4$}
\date{%
   \small{$^1$Department of Linguistics, McGill University, Montreal, Canada}\\%
   \small{$^2$Mila – Quebec Artificial Intelligence Institute, Montreal, Canada}\\%
   \small{$^3$Department of Psychology, Stanford University, Stanford, California, USA}\\
   \small{$^4$Department of Linguistics, Stanford University, Stanford, California, USA}\\[2ex]%
    January 2024
}

\begin{document}

\maketitle

\begin{abstract}
Interpreting a seemingly-simple function word like ``or", ``behind", or ``more" can require logical, numerical, and relational reasoning. How are such words learned by children? Prior acquisition theories have often relied on positing a foundation of innate knowledge. Yet recent neural-network based visual question answering models apparently can learn to use function words as part of answering questions about complex visual scenes. In this paper, we study what these models learn about function words, in the hope of better understanding how the meanings of these words can be learnt by both models and children. We show that recurrent models trained on visually grounded language learn gradient semantics for function words requiring spatial and numerical reasoning. Furthermore, we find that these models can learn the meanings of logical connectives \textit{and} and \textit{or} without any prior knowledge of logical reasoning, as well as early evidence that they are sensitive to alternative expressions when interpreting language. Finally, we show that word learning difficulty is dependent on frequency in models' input. Our findings offer proof-of-concept evidence that it is possible to learn the nuanced interpretations of function words in visually grounded context by using non-symbolic general statistical learning algorithms, without any prior knowledge of linguistic meaning.

\end{abstract}

\section{Introduction}

When studying how children learn words researchers often make the assumption that knowing the meaning of a word $w$ means having the ability to differentiate between things that are $w$ and things that are not \cite[ch.1]{bloom2002children}. This notion of meaning, sometimes called `external meaning' is in contrast to `internal meaning' -- the mental representation of meaning that a person has for $w$ -- the favored definition of meaning in theoretical semantics. Evaluating children's ability to understand the meaning of words by how they use them in the external world seems pretty straightforward in the case of nouns and predicates, but not so much for function words, like determiners, conjunctions, and prepositions. These closed-class words tend to have external meanings that only manifest themselves in how they modify other words or sentences as a whole, making them difficult to study without referring in some way to their internal meaning. Additionally, parsing their meaning often requires complex reasoning skills such as logical, numerical, spatial or relational reasoning. The abstract nature and complexity of function words are what make their acquisition by children so difficult to study using conventional methods. Yet, these same qualities are also what make function words an ideal test case to compare different theories of language acquisition and their respective learning strategies.

It has been widely observed that children tend to acquire words and grammatical structures in a specific order; this is also the case for function words. For example, \textit{and} is much more prevalent in children's linguistic input and is acquired before \textit{or} \cite{morris2008logically, jasbi2018conceptual}; Children start to correctly use the preposition \textit{behind} before they do \textit{in front of} and furthermore, their initial uses of these words are possibly conditioned on contextual factors like whether the referent object has the property of having a front and back, like a car or a doll \cite{windmiller1973relationship, kuczaj1975acquisition, clark1977strategies}, and the degree of occlusion between two objects \cite{johnston1984acquisition, grigoroglou2019pragmatics}. These differences in order of acquisition represent learning outcomes which can be used as test cases to study the impact of different types of information available in the input on learners ability to acquire these words.

Theories for the acquisition of function words tend to fall somewhere along the spectrum between nativist explanations -- for example logical nativism \cite{crain2012emergence} -- and usage-based
approaches \cite{tomasello2005constructing}. Nativist theories posit that humans are endowed with innate knowledge of some reasoning skills and that children may undergo a series of maturational stages, to reach adult like understanding. These stage-based and symbolic learning explanations predict that conceptual differences between words may lead to asymmetries in their acquisition. On the other hand, usage-based approaches argue that the reasoning skills necessary for understanding function words are learnt through experience. Children learn these words using non-symbolic general learning mechanisms which are not exclusive to language acquisition. Usage-based learning mechanisms specifically predict that frequency of exposure is a primary factor in determining the order in which new words may be learnt. While frequency may also play a role in nativist theories, it is often posited to be secondary to other conceptual differences.

\subsection{The current study}
In this paper, we will consider the acquisition of three pairs function words and their respective reasoning skills: (1) logical reasoning with the connectives \textit{or} and \textit{and}; (2) spatial reasoning with the prepositions \textit{in front of} and \textit{behind}; (3) numerical reasoning with the scalar quantifiers \textit{more} and \textit{fewer}. We hypothesise that these reasoning pairs can be learnt using non-symbolic general learning algorithms and, furthermore, that the ordering effects seen in children's acquisition of these words are simply the result of their frequency in children's input, rather than evidence for non-symbolic or stage-based learning strategies. We propose to use computational models that learn these types of words from grounded input to test these hypotheses.

We propose to use a new modeling approach which considers models as independent learners -- in other words like a new `species' of language learners -- that can be leveraged to implement `proofs of concept' (\citeNP[ch. 1.2]{lappin2021deep}, \citeNP{tsuji2021scala, warstadt2022what, pearl2023computational, portelance2023roles}).
A proof of concept can show us what is learnable `in practice' for models and `in principle' for humans. In doing so, models may be used to inform debates about the relative innateness of certain linguistic knowledge \cite{clark2011linguistic}. This approach draws on recent model interpretability work showing what kinds of grammatical knowledge language models learn  \cite{linzen2016assessing, lake2018generalization, futrell-etal-2019-neural,  manning2020emergent, Hu:et-al:2020}, and what kinds of learning biases they have (\citeNP{papadimitriou-jurafsky-2023-injecting}). With our experiments, we hope to offer proof of concept evidence showing what is in practice learnable from visually grounded language on the meaning of abstract function words requiring complex reasoning skills.

\begin{figure}
    \centering
    \includegraphics[width=\textwidth]{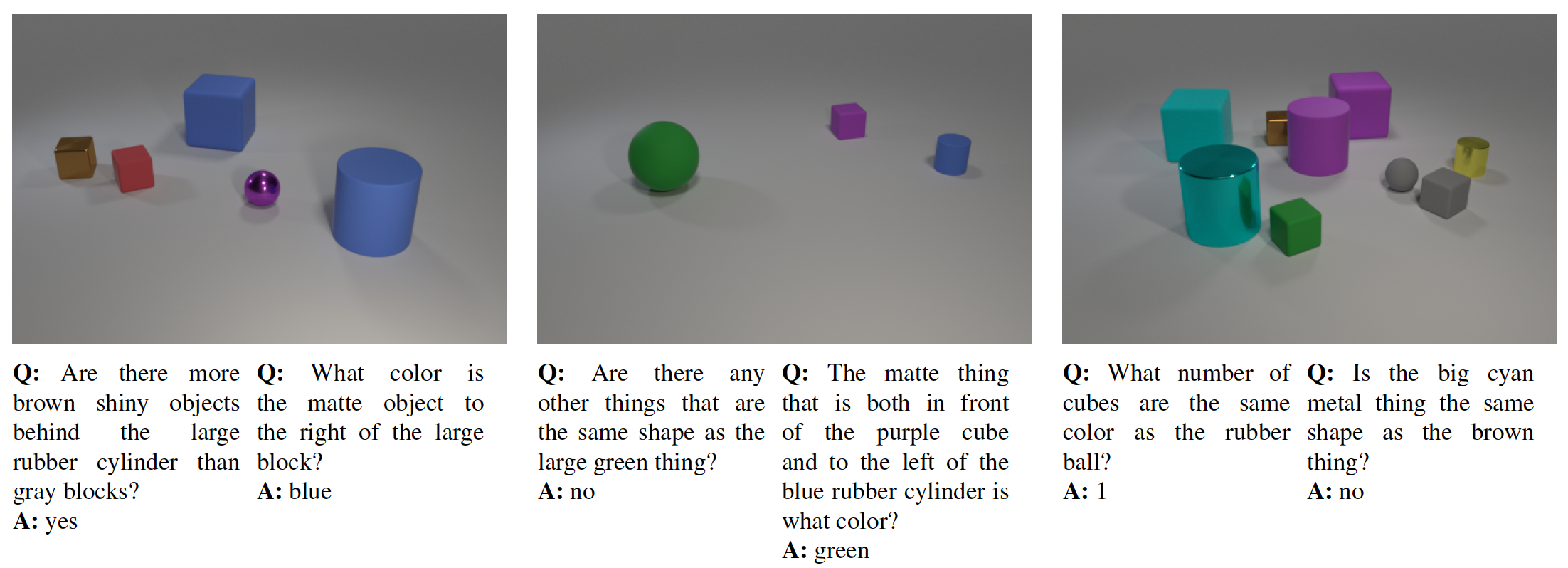}
    \caption{Example images and corresponding questions taken from CLEVR dataset.}
    \label{fig:clevr-ex}
\end{figure}

We use neural network models that learn from both linguistic and visual representations to study the effect of visual grounding on learning the meaning of function words. We can consider the interactions that may emerge from cross-modal statistical word learning, an open question developmentalists are still tackling \cite{saffran2018infant}. Specifically, we experiment with neural network models learning language in a visual question answering task, where they must come up with word representations in order to answer questions about visual scenes. The task we use is called the CLEVR (Compositional Language and Elementary Visual Reasoning) dataset \cite{johnson2017clevr}. It contains visual block-world scenes and corresponding questions like ``Are there more red cubes than metal spheres?". Models are never given the meaning of words, or any form of mapping between words and the content of images. They must deduce this information during training. Learning the meaning of words then becomes an auxiliary objective that can lead models to successfully complete their task: to generate the correct answer given some string and an image (examples from the task are given in Figure \ref{fig:clevr-ex}).

In order to propose that a neural network learner offers additional proof that some outcome -- the meaning of function words -- is likely learnable in humans, it is insufficient to just show that the models can learn this outcome; we must also weigh in on what might have led the model to learn it in the first place, and acknowledge that the proposed prerequisites for learning the outcome must also be available to human learners \cite{baroni2021proper}. Furthermore, we must outline the learning assumptions on which our proof-of-concept depends.

The learning mechanisms used by visual question answering models are almost certainly different from those used by children, but they do share one high level property: the use of indirect negative evidence. Early work suggested that children do not make use of any explicit negative evidence for word learning \cite{baker1979syntactic,pinker1989learnability, marcus1993negative, fodor2002understanding}. However,  many researchers have shown that they do rely on implicit negative evidence \cite{brown1970derivational,snow1977talking,  penner1987parental, farrar1992negative,saxton1997contrast, chouinard_clark_2003, clark2011linguistic}, for example when their desired outcomes are not met when they are misunderstood. The meanings of words may then be learnt indirectly from this evidence and the same may be said for our models. Though visual question answering models receive direct supervision on their training task -- generating correct answers to questions -- they do not receive direct supervision to learn abstract reasoning or the meanings of function words; these learning outcomes are incidental to the task and instead could be one of many strategies that models converge towards to answer the questions correctly. Our proof-of-concepts are thus conditional on the availability of some form of supervision -- direct or indirect -- during learning.

Indeed, we are not the first to make these assumptions. Visual question answering models have already been used to explore neural networks' capacity to learn meaningful representations of referential words, such as nouns and predicates, when trained on language tasks grounded in the visual world \cite{Mao2019NeuroSymbolic, pillai2021neural, zellers2021piglet, wang2021lorl, jiang2023mewl}. As for function words, \citeauthor{hill2018understanding} \citeyear{hill2018understanding} briefly consider how visually grounded models learn negation, and \citeauthor{kuhnle-copestake-2019-meaning} \citeyear{kuhnle-copestake-2019-meaning} studied how these models interpret the quantifier \textit{most}. Regier's \citeyear{regier1996human} earlier extensive work also considered how neural network models can learn to map visual scenes to spatial prepositions, though his models did not learn from any linguistic input per say and predate visual question answering models. Others more recently have also used these tasks to model noun and predicate learning in children \cite{hill2020simulating, nikolaus2021modeling}. However, to the best of our knowledge, no work has probed visually grounded neural network models' representations of the meaning of function words in the context of children's function word learning.

Throughout this paper, we will address three major research questions:

\begin{enumerate}
\item \textbf{How do visually grounded question answering models learn to represent and interpret function words and do these representations generalize to unseen linguistic and visual contexts?}
\item \textbf{Does the existence of alternative expressions in each reasoning pair affect their acquisition or are the meanings of function words acquired in isolation?}
\item \textbf{Do models learn these function words in a similar order to children and are these ordering effects the results of their frequency or do they follow from other conceptual explanations?}
\end{enumerate}

\begin{figure}
    \centering
    \includegraphics[scale=0.35]{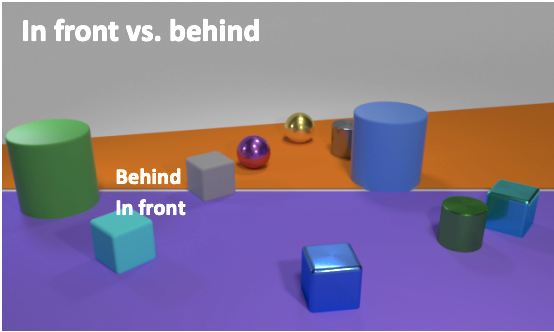}
    \caption{Threshold based interpretation of \textit{behind} and \textit{in front} relative to gray cube in CLEVR dataset. (Image from Johnson et al. 2017).}
    \label{fig:clevr-ex-behind}
\end{figure}

With respect to our first research question, each of our function words of interest are defined in absolute terms and mapped to a function over predicates in the CLEVR dataset we use. For example \textit{or} is defined as the logical operator, $A \lor B$, \textit{more} is defined as the function greater than, $|A| > |B|$, and \textit{behind} is defined as having a y-coordinate that is strictly greater than some other referent's, as in Figure \ref{fig:clevr-ex-behind}. In practice however, most of these words have much more gradient meanings when used by people in naturalistic contexts. The use of language in context distinguishes semantic representations from pragmatic interpretations. We probe how models interpret these words in novel contexts to determine how their meanings may be represented. Do their interpretations suggest that they have clear cut thresholds that distinguish the meaning of words like \textit{more} and \textit{fewer}, or does linguistic gradience arise as a result of their learning environment when exposed to grounded language use in continuous visual settings? In the CLEVR dataset, the underlying meanings of words like \textit{more} and \textit{behind} is threshold-based. So, the statement `there are more As than Bs' is always interpreted as true as long as the difference between $|A|$ and $|B|$ is over some threshold, here $|A| - |B| > 0$. Linguistic gradience on the other hand may be thought of as allowing words to have different interpretations depending on context as a function of some gradient factor. So instead, we may expect our statement `there are more As than Bs' to be interpreted as true or false as a function of the magnitude of the difference $|A| - |B|$ across contexts rather than based on some context-agnostic threshold. If models can learn representations which lead to gradient interpretations in novel contexts by using simple learning algorithms, then we can offer proof of concept evidence that these function words are learnable from supervised data using non-symbolic learning mechanisms.

With respect to our second research question, the existence of alternative expressions or worlds is the cornerstone behind Gricean pragmatic reasoning, and what allows us to have different interpretations for the same word in different contexts \cite{grice1975logic,degen2023rational}. Children have been found to exhibit pragmatic reasoning skills in multiple domains, especially when alternative worlds are made salient \cite{barner2011accessing,katsos2011pragmatic,stiller2015ad,horowitz2016children,baharloo2023children}. It is however unclear if this ability is acquired through specific means. Following Gricean theory, we might expect children to be able to judge the informativeness of contrasting expressions as soon as they have learnt their meaning \cite{katsos2011pragmatic, clark2003first}, suggesting that these abilities may stem from the same learning mechanisms. If visual question answering models can learn to consider alternative expressions when interpreting function words like \textit{and} and \textit{or} in novel contexts, then we may offer proof of concept evidence that the ability to reason about alternatives can be derived from statistical learning mechanism applied in a contextually grounded setting.

With respect to our third research question, frequency or word predictability is a known predictor of the order in which children acquire words \cite{goodman2008does, kuperman2012age,braginsky2019consistency, portelance2023predicting}.
There may however be other factors -- in relation with or independent from -- frequency that make learning the meaning of certain function words harder than others. For example, Clark \citeyear{clark_1993mapping} points out that there seems to be an asymmetry in the acquisition of adjective pairs like \textit{big} and \textit{little}, \textit{tall} and \textit{short}, etc., where children tend to produce words for positive dimensions before they do negative ones. This difference in learning may be independent from frequency, since in experiments where children are exposed to nonsense word pairs like these with even frequency, they still seem to favor learning the positive words over the negative ones \cite{klatzky1973asymmetries}. These results would then promote a conceptual explanation for these effects over a frequency-based explanation. Such asymmetries may also exist for similarly polarized pairs of function words. Here, we explore whether the order in which these words are learnt is a function of how frequent they are in the input or if there may be other factors which makes certain function words intrinsically more difficult to learn than others. We will compare the order in which children acquire words requiring similar abstract reasoning to the order in which visual question answering models learn these same words, while varying their relative frequency in the models' input.

Our approach is as follows. We define a novel semantic testing task within the CLEVR block world to determine whether models understand the meanings of function words in unseen contexts. We then evaluate model performance on these novel tests throughout training to visualize how learning progresses. Next, we compare the relative order in which models learn our function words to the acquisition order we expect in children. We manipulate input distributions and train models on different subsets of the training data with various function word frequencies to analyse whether the ordering effects initially observed are solely mediated by frequency or if other more conceptual factors play a role. \footnote{All of the data, models, and experiment code presented in this paper are publicly available at \href{https://github.com/evaportelance/vqa-function-word-learning}{\texttt{www.github.com/evaportelance/vqa-function-word-learning}}.}
In the remainder of this introduction, we briefly review children's acquisition of function words and the visual question answering task and dataset we use.

\subsection{Children's acquisition of the target function words}

For each of the word pairs and their respective reasoning skills considered in this study (``and"/``or", ``behind"/``in front of", ``more"/``fewer"), we review what is currently known and debated about their acquisition in the child language learning literature. We note that most of the previous research on these words is exclusively about English, with a couple exceptions, mentioned when relevant.

\subsubsection{``and'' / ``or''}

The source of the emergence of logical reasoning in children has been debated for quite some time (for a thorough review of the field see \citeNP[Ch.5]{jasbi2018learning}). Proposals tend to fall somewhere along the spectrum between logical nativism \cite{crain2012emergence} and usage-based approaches \cite{morris2008logically}. Logical nativism posits that humans are endowed with innate logic and children then go through a series of developmental stages to reach adult like logical understanding. As for usage-based approaches, these argue that logical reasoning is learnt through experience using general learning mechanisms-- as opposed to learning strategies that are specific to logical reasoning -- and that frequency in children's input explains any ordering effects seen in children's learning of logical concepts.

All agree that children correctly interpret \textit{and} before \textit{or}; \textit{and} is also much more frequent than \textit{or} in children’s input, and furthermore, they are exposed to more instances of exclusive \textit{or} than inclusive \textit{or} \cite{morris2008logically, jasbi2018conceptual}. There is however some debate about the order in which children acquire possible meanings of \textit{or} and what the underlying meaning of this logical connective may be in children's representations. Given its higher frequency, \citeauthor{morris2008logically} \citeyear{morris2008logically} suggests that children initially learn exclusive \textit{or}. Similarly, early nativist approaches argued that children's early understanding of \textit{or} was as a simple choice, making it compatible with exclusivity \cite{neimark1970development}. Following Grice's \citeyear{grice1975logic} proposal that exclusive interpretations are the result of generalized conversational implicature, others have instead advocated that \textit{or} is underlyingly inclusive and that children eventually learn exclusive \textit{or} via pragmatic reasoning \cite{chierchia2001acquisition, chierchia2004semantic, jasbi2021adults}. Interestingly, some have also found the children often mistakenly interpret \textit{or} as conjunction \cite{braine1981development, singh2016children, tieu2017role}, though it has been suggested that this finding may be an artifact to the specific experimental task designs used in these studies \cite{paris1973comprehension,skordos2020do}.

All of the experimental results showing that children understand \textit{or} inclusively still leave unanswered the question of how they come to learn the meaning of this word in the first place. Crain \citeyear{crain2008interpretation, crain2012emergence} argues that these results are in fact evidence in favor of a logical nativist explanation since, though children are exposed to more instances of exclusive interpretations of \textit{or}, they seem to instead favor inclusive interpretations initially. Currently, there is little evidence showing that inclusive \textit{or} is learnable from more general learning mechanisms that would support a usage-based approach.

\subsubsection{``behind'' / ``in front of''}

Children learn the meaning of the locative preposition \textit{behind} before they do \textit{in front of} \cite{johnston1979development, johnston1984acquisition}. There have been a few proposals for explaining this asymmetry, all sharing a common thread: that children do not initially encode the meaning of these words in geometric spatial terms. The semantic mis-analysis hypothesis for the asymmetry in children's early understanding of these expressions suggests that children struggle to incorporate the perspective of the observer in analysing the meanings of these words \cite{piaget1967child}, so they erroneously define the concepts of \textit{front} and \textit{back} in terms of visibility and occlusion \cite{johnston1984acquisition}. \citeauthor{grigoroglou2019pragmatics} \citeyear{grigoroglou2019pragmatics}, also suggest that children analyse these words in terms of occlusion but not as a result of semantic misanalysis, instead as the result of pragmatic inference, where occlusion is more notable than visibility. Much of the research on the acquisition of \textit{behind} and \textit{in front of} then documents the stages of development between these early word representations and their adult-like geometric meanings. They conducted experiments in both English and Greek. Some researchers have found that this transition is aided by the eventual projection of the property of having a front or back on objects (e.g. being behind a doll versus being behind a block) \cite{windmiller1973relationship, kuczaj1975acquisition, clark1977strategies}. Again, there is currently a lack of evidence supporting the use of more general learning mechanisms behind the acquisition of these words, as opposed to learning strategies specific to spacial reasoning.

\subsubsection{``more'' / ``fewer"}

Quantifiers have been found to follow quite robust acquisition ordering effects cross-linguistically \cite[analysis over 30 languages]{katsos2016cross}. For the comparative quantifiers \textit{more (than)} and \textit{fewer (than)}, the meaning of \textit{more} has repeatedly been found to be learnt  earlier than \textit{fewer/less} by children \cite{donaldson1968less, palermo1973more, donaldson1970acquisition, townsend1974children, geurts2010scalar}. Some have also found that children initially interpret \textit{less} as a synonym of \textit{more} \cite{donaldson1968less, palermo1973more}, but as \citeauthor{townsend1974children} \citeyear{townsend1974children} points out, these earlier experimental studies did not have a way to truly distinguish between children interpreting \textit{less} as \textit{more} or simply not knowing the meaning of \textit{less}. A few hypotheses have been put forward to explain the acquisition asymmetry between these two comparative quantifiers, all favoring conceptual explanations over frequency-based ones. Though \citeauthor{donaldson1970acquisition} \citeyear{donaldson1970acquisition} briefly mention that \textit{more} is much more frequent than \textit{less} in children's input, they quickly reject the possibility that frequency is the answer, arguing that if the asymmetry was down to frequency, we would expect children that do not know the meaning of \textit{less} to interpret this word in a variety of ways. However, citing previous work, they suggest that \textit{less} is instead always interpreted as \textit{more}. They thus propose that there are a series of developmental stages for the processing of comparatives, which lead to this asymmetry, where \textit{more} is acquired earlier because children initially learn to use it in singular referent contexts like in the additive sense of \textit{more}, for which they say a counterpart with \textit{less} is not possible. \citeauthor{clark1970primitive} \citeyear{clark1970primitive} offers a similar proposal with slightly different developmental stages. Still, these results clearly suggest that word frequency might account for   developmental ordering phenomena, consistent with usage-based accounts as well.

\section{Evaluating function word knowledge using semantic probes}

As a testbed for the learnability of function words, we use visual question answering models trained on the CLEVR dataset, a standard dataset used in the broader natural language processing community \cite{johnson2017clevr}. We propose a semantic probe zero-shot evaluation task based on CLEVR to determine whether models were able to learn meaningful representations for each of reasoning pairs under study: \textit{and}/\textit{or}, \textit{behind}/\textit{in front of}, and \textit{more}/\textit{fewer}\footnote{In appendix C we also include some experiments with relational reasoning and the adjective \textit{same}.}.

\subsection{Visual question answering and the CLEVR dataset}

Visual question answering was proposed as a language learning task that is grounded in images and requires models to develop abstract reasoning skills \cite{malinowski2014multi, antol2015vqa, gao2015you, ren2015exploring}. Models are given images and questions about their content as input; they are then trained to answer these visually grounded questions (example image-question pairs from the CLEVR dataset are given in Figure \ref{fig:clevr-ex}). Generating the correct answers often requires reasoning skills, such as logical reasoning, spatial reasoning, and numerical reasoning, which models must also learn. Since learning the meaning of function words requires developing these same reasoning skills, models trained to complete these types of tasks lend themselves well to the study of function word learning using neural networks.

Initial visual question answering tasks used datasets that were produced by having human annotators come up with questions for images \cite{malinowski2014multi, antol2015vqa, gao2015you, krishna2016visualgenome}. However, as the first resulting models emerged it became clear that they had shortcomings which prevented them from developing abstract reasoning, in part due to unbalanced datasets \cite{agrawal-etal-2016-analyzing, zhang2016yin}. To avoid this problem and to help parse which reasoning skills models were developing and relying on, balanced datasets with explicit generative models to produce questions \cite{johnson2017clevr, hudson2019gqa} and images \cite{johnson2017clevr} were created. CLEVR is one such dataset, containing generated images of scenes from a 3D block-world and constructed questions.

We chose this dataset as it offered us the benefit of precisely defining the function words in the dataset by associating them to explicit functional relations, giving us a better grasp over the underlying semantics of these words. For these reasons, the CLEVR dataset \cite{johnson2017clevr} serves as a good starting point for our comparison between between VQA model and children's acquisition of function words. Specifically, as mentioned in the introduction, \textit{or} is defined as the inclusive logical operator, $A \lor B$, while \textit{and} is $A \land B$; \textit{more} is defined as greater than, $|A| > |B|$, while \textit{fewer} is $|A| < |B|$; and \textit{behind} is defined as having a y-coordinate that is strictly greater than some other referent's, $y(a) > y(b)$, while \textit{in front} does the opposite $y(a) < y(b)$, as in Figure \ref{fig:clevr-ex-behind}. Additionally, it is a well-balanced dataset whose composition has been extensively described and well understood \cite{johnson2017clevr}.

It is composed of questions paired with images like those illustrated in Figure \ref{fig:clevr-ex}. The images are all of complex scenes in a block-world involving static objects placed on a 3D grey plane. Objects have four varying attributes: shape, color, material, and size. The number of objects in an image varies randomly between 3 and 10, as does their relative positions and the positions of light sources in the scenes. There are a total of 70,000 distinct images in the training set and another 15,000 different images in the validation set.\footnote{The CLEVR dataset also contains a test set, but since this dataset was designed  as a benchmarking task, the meta-information for test images isn't publicly available, nor are the answers to the test questions. We tried contacting the authors of the original paper to gain access to the test images' meta-information in order to use them for our probe design, but we were unsuccessful. For these reasons, the images from the validation set were used in designing our semantic probe testing task.}

Each image is paired with a set of questions like those in Figure \ref{fig:clevr-ex}. In total there are 699,989 questions in the training set and 149,991 in the validation set. There are different types of questions, including existential questions, count questions, attribute identification questions, and comparison questions, requiring a slew of reasoning skills to answer them. Questions can be compositional and require multiple reasoning steps to arrive at the right answer. For a break down of all the question types and a full definition of the generative model used to generate them, we refer the reader to the original CLEVR dataset paper \cite{johnson2017clevr}.

The CLEVR dataset is a standardized and highly-controlled dataset intended to facilitate progress in the development of natural language processing systems, but it is not natural language; it does not have all the same properties as the speech children are exposed to. The language in CLEVR is template-based\footnote{Examples of the templates in question containing our function words are given in Appendix A.} and text only; by contrast, children's input is composed of a much richer signal including varied syntactic frames, prosody, social cues, and other sources of information. This fundamental difference means that our models do not have access to much of the rich information that children leverage to learn new words. On the other hand, natural environments are also noisier; a constrained learning environment may inadvertently help models learn and converge on the tasks quicker. Working within a highly controlled and simplified learning environment is a necessary first step to understand the relations that exist between models' input and their learning outcomes.

\subsection{Semantic probes}

\begin{figure}
    \centering
    \includegraphics[width=\textwidth]{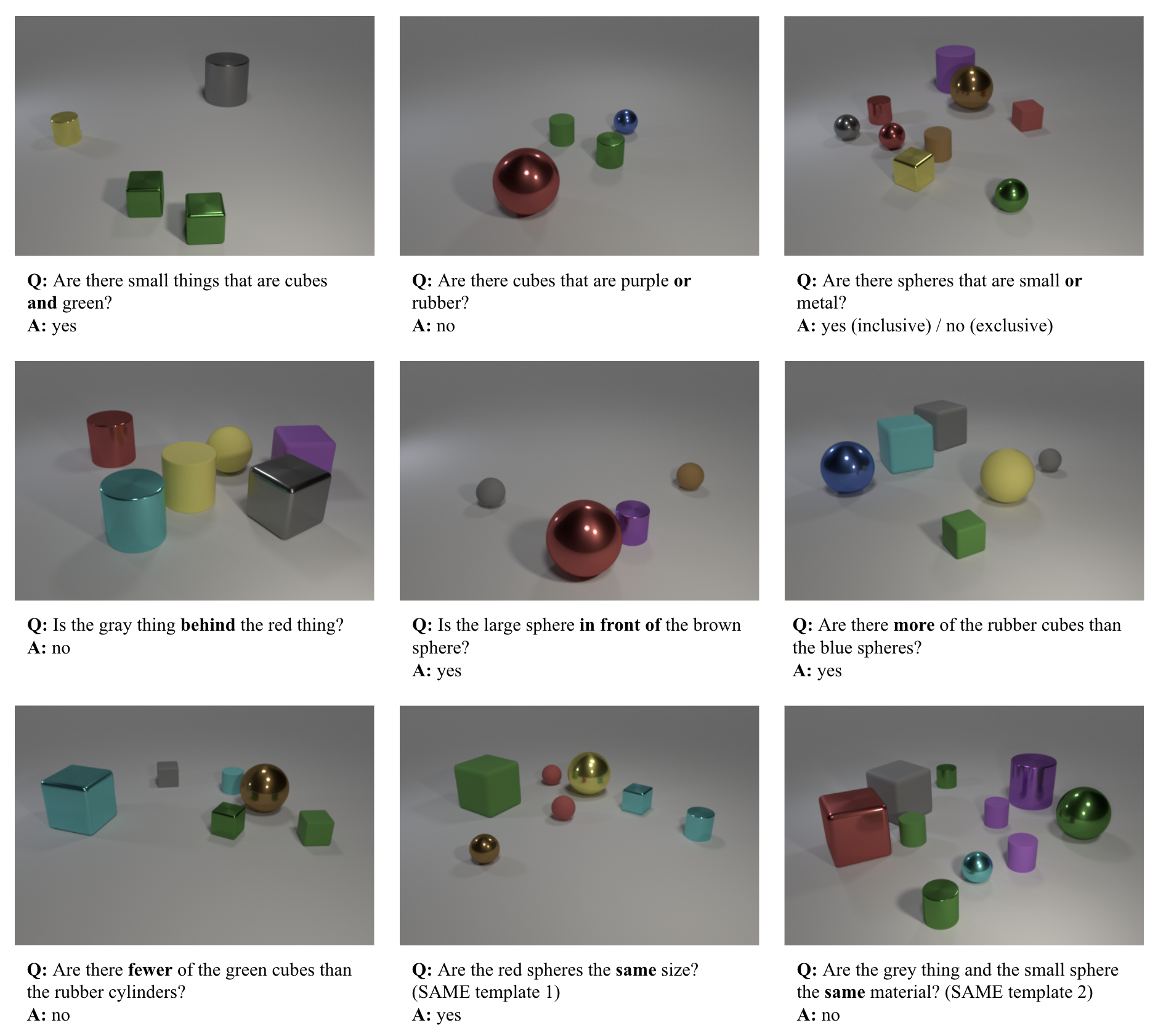}
    \caption{Example image-question pairs from semantic probes.}
    \label{fig:probe-ex}
\end{figure}

Each semantic probe is a set of existential questions based on a simple template that contains one of our function words of interest. Models must know the meaning of the relevant word to answer probe questions correctly, otherwise, we would expect performance to be at or below chance on probe questions overall. Each question is associated with an image from the CLEVR validation image set that satisfies any implied presuppositions. Example image-question pairs from each probe are presented in Figure \ref{fig:probe-ex}. The probes are all based on unseen templates, though they are all composed of words which are part of the CLEVR vocabulary and show some similarities with existing CLEVR question templates.

\begin{figure}
    \centering
    \includegraphics[width=\textwidth]{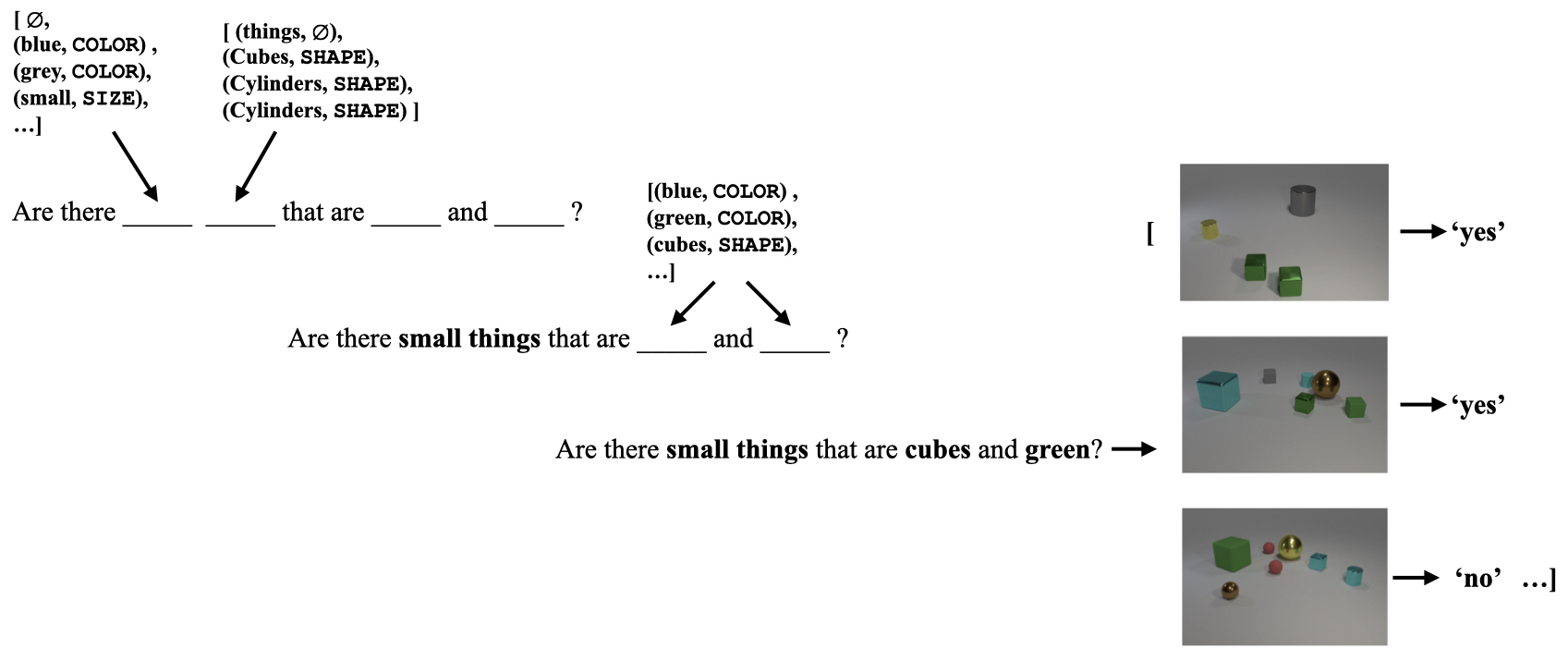}
    \caption{Example probe creation procedure. Given a template, we cycle through every possible variable combination and then sample 10 images and determine their corresponding answers.}
    \label{fig:probe-deriv}
\end{figure}

For each probe, given the template, we created the set of questions such that we iterate through every possible combination of referents in the CLEVR universe, allowing us to abstract away any difficulty answering questions that may be due to other content words. For each question, we then identified all the images in the validation set that met its presuppositions. If there were more than 10 such images we randomly sampled 10 of them. Figure \ref{fig:probe-deriv} illustrates this procedure. In the rest of this paper, we will use the capitalized version of a word to refer to its respective semantic probes, for example, AND will refer to the semantic probe for the word \textit{and}.

\paragraph{AND - OR}
probes templates are `Are there $X$s that are $\alpha$ \textbf{and} $\beta$?' and `Are there $X$s that are $\alpha$ \textbf{or} $\beta$?', where $X$ is a referential expression (e.g. gray sphere, metal thing, big cylinder, cube) and $\alpha, \beta$ are properties (e.g. purple, small, metal). As previously mentioned, the probes iterate through every possible referent combination, where a referent is noun (thing, sphere, cylinder, cube) optionally preceded by a modifier referring to its color, material or size. These templates do not have any presuppositions, so 10 images were randomly sampled for each one, totalling 15,600 image-question pairs in each probe.

For the AND probe, questions which were paired with images that contained at least one $X$ that was both $\alpha$ and $\beta$ -- $(\alpha \land \beta)$ -- had `yes' as their correct answer, while questions where this requirement wasn't met in the image had `no' as their answer.

We were interested in determining the prevalence of inclusive versus exclusive interpretations of the word \textit{or} by models. For this reason, we used the following answer scheme for the OR probe. Questions which were paired with images that contained at least one $X$ that was $\alpha$ but not $\beta$ -- $(\alpha \land \neg \beta)$ --, or not $\alpha$ but $\beta$ -- $(\neg \alpha \land \beta)$ --, expected the correct answer `yes'. Questions which were paired with images that contained no $X$s or only $X$s that were neither $\alpha$ nor $\beta$ -- $(\neg \alpha \land \neg \beta)$ -- had `no' as there answer. As for question-image pairs where all $X$s were both $\alpha$ and $\beta$ -- $(\alpha \land \beta)$ -- were ambiguous, expecting a `yes' answer if \textit{or} was interpreted as inclusive, while a `no' answer if on the other hand \textit{or} was interpreted as exclusive.

\paragraph{BEHIND - IN FRONT OF} probes used as templates `Is the $X$ \textbf{behind} the $Y$?' and `Is the $X$ \textbf{in front of} the $Y$?', where both $X$ and $Y$ are referential expressions. These templates presuppose that the images contain exactly one $X$ and one $Y$. Again iterating over the same complete set of referent combinations\footnote{There is an exception for the noun `thing' in the case of BEHIND - IN FRONT OF and MORE - FEWER probe templates which obligatorily requires a modifier, e.g. 'Is the \textit{blue} thing behind the sphere?' We must include some modifier like \textit{blue} or the referent cannot be uniquely identified.}, we identified all the images which satisfied this presupposition. If there were more than 10, we randomly sampled 10 of them, otherwise we included all available images. In the end there were a total of 24,380 image-question pairs for each probe.

Using the `scene' metadata available for each image, which contains annotations as to the relative position of objects, we determined the correct answer to each question. These relative positions were determined using the $(x,y,z)$ center point coordinates of objects. Using the underlying threshold definitions of \textit{behind} and \textit{in front} from CLEVR, we determined if an object was behind or in front of another by taking the difference between their $y$ coordinates. Image-question pairs where $X$ was in fact behind $Y$ received a 'yes' answer for the BEHIND probe and a `no' answer for the IN FRONT OF probe. If the opposite was true, the answers were reversed. In our analyses, we additionally wanted to track probe questions performance based on the relative distance between $X$ and $Y$. For these analyses we kept track of the Euclidean distance between the two referent objects using all three of their $(x,y,z)$ coordinates.

\paragraph{MORE - FEWER} probes follow the forms `Are there \textbf{more} of the $X$s than the $Y$s?' and `Are there \textbf{fewer} of the $X$s than the $Y$s?'. Both these templates presuppose that the images contain at least one $X$ and one $Y$. Based on this presupposition we identified all of the compatible images for each question and, again, if more than 10 images were found we randomly sampled 10 of them for a given question. In total there were 24,420 image-questions pairs in each of these probes.

To determine the answers to each image-question pair, we once again used the `scene' metadata which was associated to each image. We identified all of the objects which were part of $X$ and $Y$ referent categories and then compared their cardinality. {Based on our underlying CLEVR definitions,} if the number of $X$s was greater than the number of $Y$s, ($|X| > |Y|$), then the answer to a question in the MORE probe was `yes', while the answer to a question in the FEWER probe was `no'. If on the other hand the number of $X$s was less than the number of $Y$s, ($|X| < |Y|$), then the opposite answering pattern applied, MORE questions had `no' for an answer, while FEWER questions - `yes'. In the event that there were the exact same number of $X$s and $Y$s, ($|X| = |Y|$), both probe question types' answer was `no'. We were interested in tracking model performance on probe questions as a function of the difference in cardinality between the two referent sets, ($|X| - |Y|$), so we also kept track of this number for each image-question pair.

\subsection{Evaluation} \label{sec:eval}
In each of the experiments that follow, we use these probes to evaluate how much models have learnt about the meaning of these words and how they interpret them given different visual contexts. We test models on all probes at each epoch during model training, allowing us to analyse what they are learning over time. As we do these analyses, it is important to understand certain distributional facts about the training data our models are exposed to.

\begin{table}
    \centering
    \begin{tabular}{|l|c|c|}
         \hline
         word pairs & raw counts & frequency\\
         \hline
         and & 81,506 & 56.32\% \\
         or & 63,214 & 43.68\% \\
         \hline
         behind & 147,409 & 49.98\% \\
         in front of & 147,506 & 50.02\% \\
         \hline
         more & 11,570 & 49.40 \% \\
         fewer & 11,851 & 50.60 \% \\
         \hline
    \end{tabular}
    \caption{Relative frequencies of each function word pair in the CLEVR training data.}
    \label{tab:clevr-freq}
\end{table}

\begin{table}
    \centering
    \begin{tabular}{|l|c|c|c|c|}
    \hline
    & \multicolumn{2}{c |}{\textit{yes} answers} & \multicolumn{2}{c |}{\textit{no} answers}\\
word pair & raw counts & frequency  & raw counts & frequency\\
         \hline
         and & 20,673 & 25.36 \% & 21,463 & 26.33 \%\\
         or & 0 & 0\% & 0 & 0\%\\
         \hline
         behind & 27,491 & 18.65\% & 28,707 & 19.47\%\\
         in front of & 27,748 & 18.81\% & 28,563 & 19.36\%\\
         \hline
         more & 5,549 & 47.96 \% & 6,021 & 52.04 \%\\
         fewer & 5,840 & 49.28 \% & 6,011 & 50.72 \%\\
         \hline
    \end{tabular}
    \caption{Frequencies of \textit{yes} and \textit{no} answers for questions containing each function word in the CLEVR training data.}
    \label{tab:clevr-ans-freq}
\end{table}

The CLEVR dataset is well balanced in terms of the relative frequency of each function word. Table \ref{tab:clevr-freq} shows the raw counts for words as well as their relative frequency by word pair in the training data. The total number of word tokens is 12,868,670 words, over 699,989 training questions.

Additionally, `yes' and `no' answers to questions containing these words are also generally well balanced, the exception being questions containing the word \textit{or}. Table \ref{tab:clevr-ans-freq} shows the relative frequencies of these answers for questions containing each of our function words. As evident from this table, there are no questions containing the word \textit{or} which are answered using `yes' or `no'.  \textit{Or} is always used as a logical conjunct connecting referents, specifically in count questions (e.g. `How many things are blue cubes or small cylinders?'), which all require a number as their answer. All the while, \textit{and} is additionally used in a much wider variety of question types, sometimes connecting prepositional phrases (e.g. `What material is the blue cube that is behind the cylinder and left of the red thing?'). Cumulatively, about 52 \% of questions with \textit{and} require a yes/no answer, while the rest are other words in the vocabulary. Like \textit{and}, \textit{behind} and \textit{in front of} show up in a variety of question types, requiring different types of answers, while \textit{more} and \textit{fewer} are only used in questions which require `yes' or `no' answers. These differences in input distributions are artifacts of the CLEVR dataset generator and the question templates used by the original authors behind this dataset. Thus, in the results which follow, it is difficult to fairly compare across word pairs or across AND and OR probes; we should instead consider them somewhat independently. However, if we observe differences in results within well balanced pairs, these are likely due to other factors beyond their frequency in the models' input. We will explore some of these factors further in the experiments that follow.

\section{MAC: A recurrent reasoning model for question answering}
\begin{figure}
    \centering
    \includegraphics[width=\textwidth]{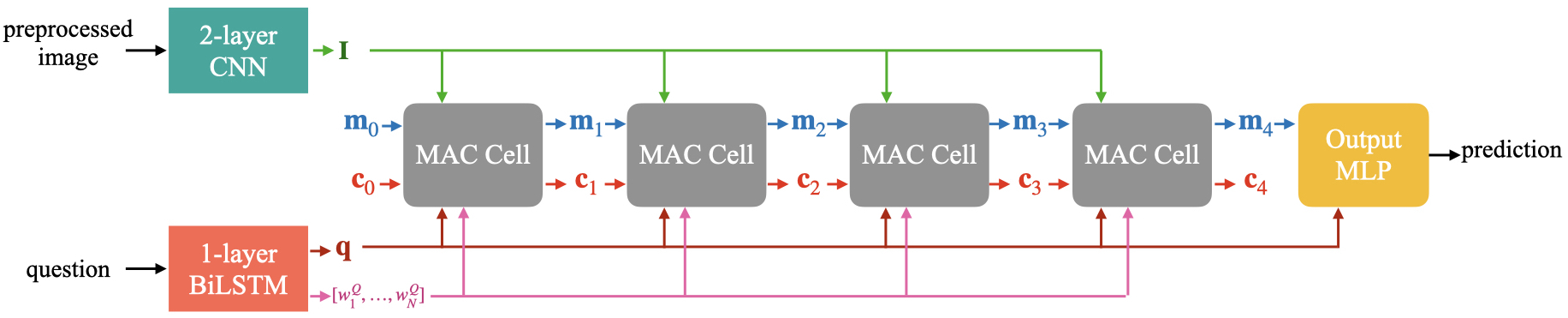}
    \caption[The MAC model with recurrent processing states.]{The MAC model initially processes the image and question through CNN and biLSTM units respectively and then through four recurrent MAC cells, each generating output memory and control states. The final output unit takes the final memory state and the question representation to produce a prediction. The model is fully differentiable.}
    \label{fig:mac-model}
\end{figure}

A variety of models have been proposed for completing visual question answering tasks; all of these include both visual and linguistic processing units. For our current experiments, we chose to use a model that -- at the time we began the project -- had the top performance scores on the original CLEVR task: the MAC (Memory, Attention, and Composition) model of \citeNP{hudson2018compositional}. This model reaches an accuracy level of 98.9\% on CLEVR's test set. Because it does so well on within-sample questions, we hoped that it could also generalize to our probe questions as well.

Following previous approaches to the CLEVR task \cite{hu2017learning, santoro2017simple, perez2018film}, the MAC model preprocesses images using ResNet-101 \cite{he2016deep}, pre-trained on ImageNet \cite{russakovsky2015imagenet}. The \textit{conv4} layer features from ResNet-101 are then used to represent each image.

The MAC model is a recurrent reasoning model which we illustrate in Figure \ref{fig:mac-model} and describe in what follows. It first processes the preprocessed image and question separately.  The preprocessed image goes through a two layered convolutional neural network resulting in a three dimensional matrix (preprocessed image width $\times$ preprocessed image height $\times$ number of channels in final convolutional layer) representing what Hudson and Manning call \textit{the knowledge base}, $\bm{I}$. As for the question $Q$, each word is converted to an embedding vector and then processed through a single layered bidirectional long-short term memory (biLSTM) network. The biLSTM yields two outputs for a question $Q$ of length $N$ words: (1) a vector of contextualised word embeddings $[w^Q_1, \ldots, w^Q_N]$, where each $w^Q_n$ is the models output state for $w_n$; (2) a question representation $\bm{q}$ which is the concatenation of the final states of both the forward and backward passes of the biLSTM, $\bm{q}=[\overleftarrow{w^Q_1}, \overrightarrow{w^Q_N}]$. Once the image and question are processed as $\bm{I}$, $[w^Q_1, \ldots, w^Q_N]$, and $\bm{q}$, they are used as input for a set of recurrent reasoning steps.

The MAC model uses custom recurrent cells (MAC cells) which each represent one reasoning step $t$. The best version of the MAC model as originally reported used 12 recurrent MAC cells before the final output layer. Hudson and Manning however found that very similar performance could be achieved with as few as 4 recurrent reasoning steps (test accuracy 97.9\%). Thus, we chose to use this smaller and more efficient version of the model for our experiments -- see Figure \ref{fig:mac-model} for a visualization of our version of the model\footnote{We kept all other hyperparameters the same as the ones used in the main version of the MAC model in \citeNP{hudson2018compositional} (see appendix B).}. Each reasoning step $t$ between 1 and 4, the MAC cell takes as input the processed image representation $\bm{I}$, the contextualized word embeddings $[w^Q_1, \ldots, w^Q_N]$, and the processed question representation $\bm{q}$. Additionally, as these are recurrent cells, it also considers two hidden states as input: (1) one representing a soft attention map over the question, $\bm{c_{(t-1)}}$ (called the control state in \citeNP{hudson2018compositional}); (2) the other representing soft attention map over the image, $\bm{m_{(t-1)}}$  (called the memory state) , where $\bm{c}_{0}, \bm{m}_0$ would be randomly initialized dummy vectors. The output of a recurrent cell at reasoning step $t$ is then $\bm{c}_{t}$ and $\bm{m}_t$, which can then be used as the hidden states for the next reasoning step. At the final step 4, the model integrates the final soft attention map over the image representation $\bm{m}_4$ with the question representation $\bm{q}$ through a basic multi-layer perceptron (MLP) to predict an answer, which always consist of a single word from the model's shared question and answer vocabulary.

\begin{figure}
    \centering
    \includegraphics[width=\textwidth]{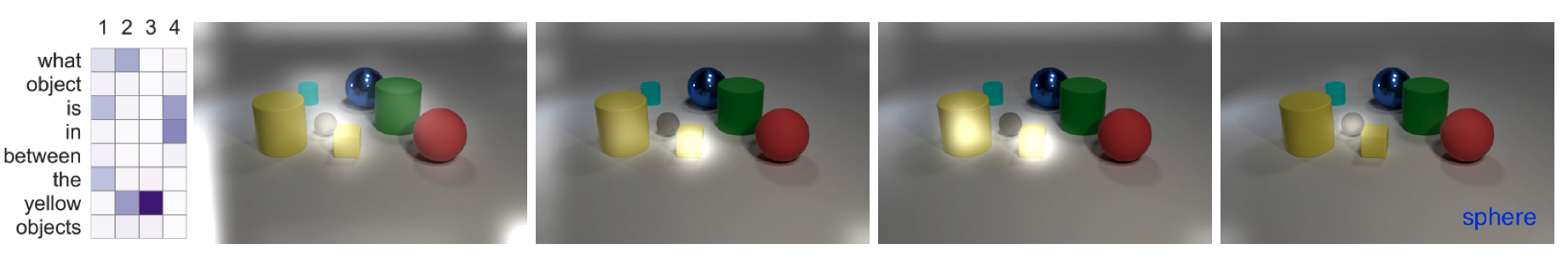}
    \caption{Example attention maps produced
by MAC model at each reasoning step taken from Hudson \& Manning, 2018.}
    \label{fig:mac-deriv}
\end{figure}

The control state $\bm{c}_t$ is a weighted distribution over the contextualized word embeddings. In other words, it indicates which words are most important to attend to in a given reasoning step. The memory state $\bm{m}_t$ is a weighted distribution over regions in the processed image which is conditioned on $\bm{c}_t$. Intuitively, it encodes which parts of the image to attend to given the parts of the question being considered at a given reasoning step. Example outputs of both memory and control states 1 through 4 for a given question image pair can be seen in Figure \ref{fig:mac-deriv}.

For a detailed breakdown of the MAC cell's internal structure and how these attention maps are derived we refer the reader to \citeNP{hudson2018compositional}. For the purpose of this paper we note that the cell has a relatively simple and straightforward structure composed of separate MLPs for processing the control $\bm{c}_{t}$ and $\bm{m}_{t}$ memory states. It was designed ``to capture the inner workings of an elementary, yet general-purpose reasoning step" and to ``encourage the network to solve problems by decomposing them into a sequence of attention-based reasoning operations that are directly inferred from the data, without resorting to any strong supervision" \cite{hudson2018compositional}. The model's generic and simple structure eliminate the possibility of it introducing any form of symbolic structural biases, which is important since it will serve as an example of non-symbolic learning for our hypotheses testing.

\section{Experiment 1: Learning to interpret and represent function words}

How do visually grounded question answering models learn to represent and interpret function words? Do the representations they learn for words like \textit{and}, \textit{or}, \textit{behind}, \textit{in front of}, \textit{more}, and \textit{fewer} generalize to unseen linguistic and visual contexts?

\subsection{Setup}

We trained five MAC models on the original CLEVR training data for 25 epochs, initialized using different random seeds. Models learn and update using back propagation with the addition of variational dropout on 15\% of parameters across the model at each pass. Models reached an average prediction accuracy of 98.84\% on the training data and of 97.74\% on the validation set, reproducing the performances originally reported by \citeauthor{hudson2018compositional} (2018) for 4-step MAC models. Since our probes are based on never seen question templates, we expect models' performance on probes to be lower than their performance on the CLEVR's validation set which was created using the same question templates as the training data. We report mean F1 score and standard deviation across all five models for each probe at each epoch throughout training. Chance performance is in theory a near 0 F1 score, since models can produce any word in their vocabulary as the answer to probe questions. However, models very quickly learn after only a couple batches that existential questions are always answered with either `yes' or `no', significantly reducing the number of answers they actually consider.

\subsection{Results}

\begin{figure}[H]
    \centering
    \includegraphics[height=6cm]{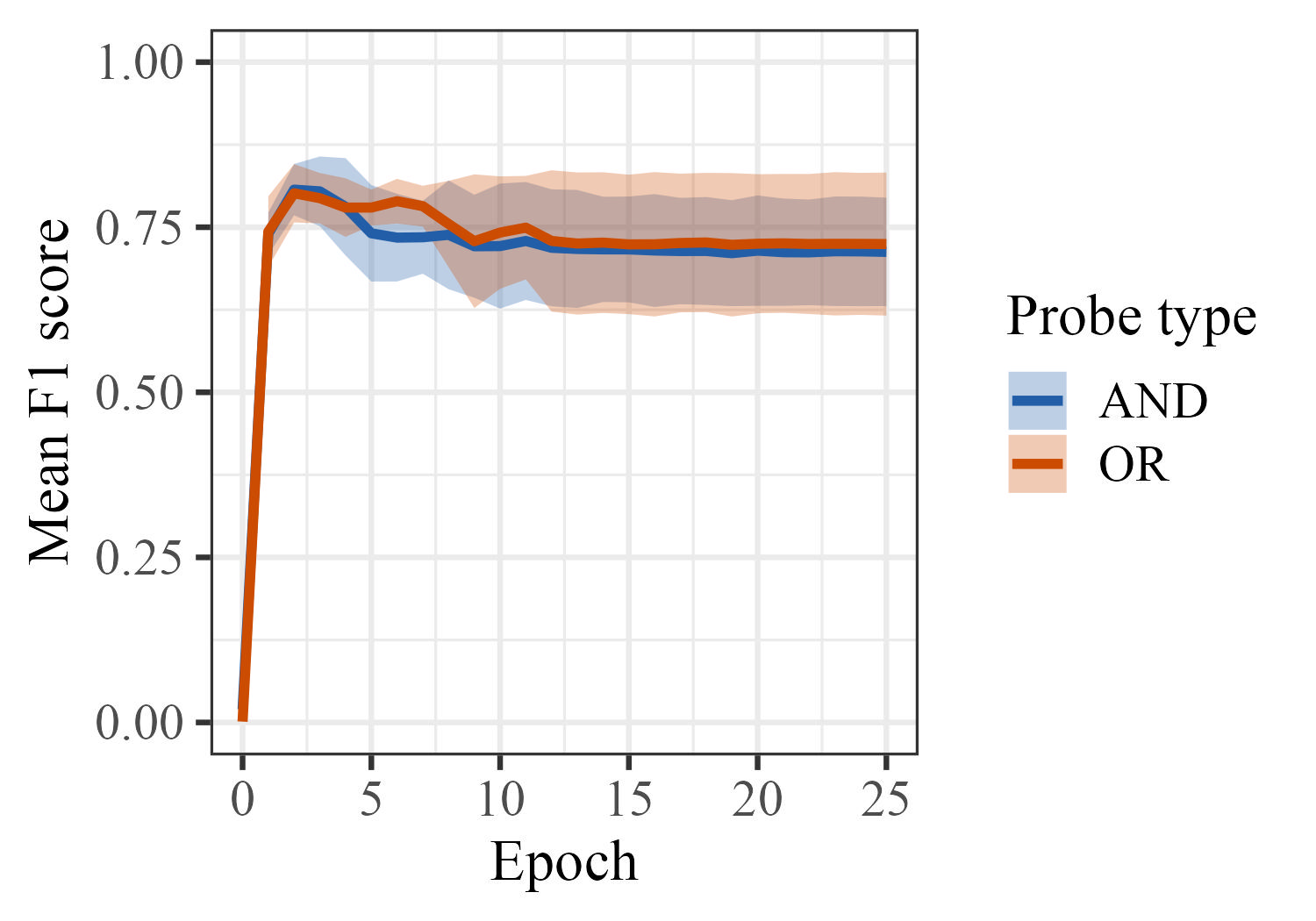}
    \caption[Experiment 1: F1 score on AND - OR probes overall.]{Experiment 1: Mean F1 score on AND - OR probes overall in non-ambiguous questions, shading represents standard deviation across 5 models.}
    \label{fig:andor-over}
\end{figure}

\paragraph{AND - OR} Probe questions were all of the form `Are there $X$s that are $\alpha$ and/or $\beta$?'. As a reminder, there are four possible truth-conditions associated to the images the questions are paired with: $(\alpha \land \beta)$, $(\alpha \land \neg \beta)$, $(\neg \alpha \land \beta)$, and $(\neg \alpha \land \neg \beta)$. First, let's consider the overall scores of models on probes in non-ambiguous contexts in Figure \ref{fig:andor-over} -- in other words, excluding OR probe questions in $(\alpha \land \beta)$ contexts, where inclusive and exclusive interpretations of \textit{or} have opposing answers. As seen in the figure, models perform better than chance on both the AND and OR probes.

\begin{figure}[H]
    \centering
    \includegraphics[height=6cm]{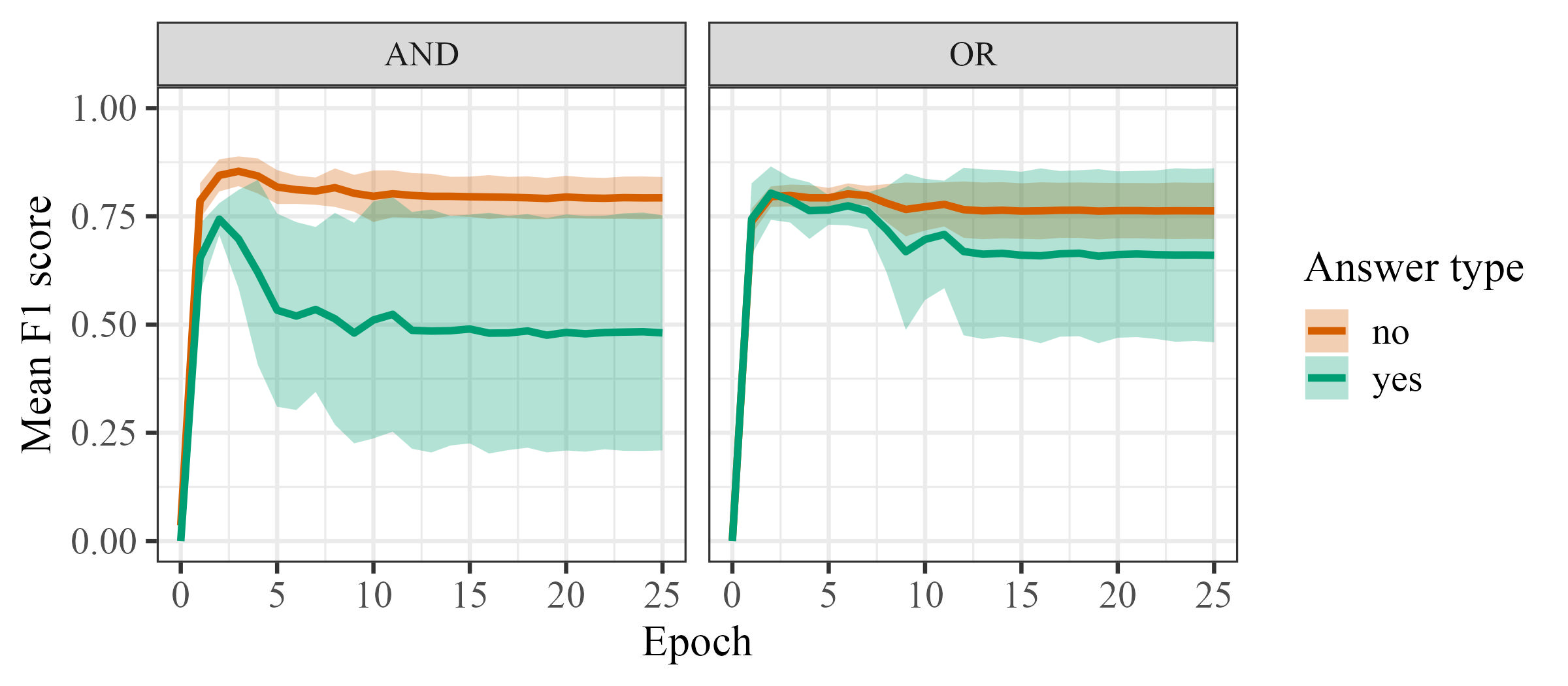}
    \caption[Experiment 1: Mean F1 score on AND - OR probes by answer type.]{Experiment 1: Mean F1 score on AND - OR probes by answer type in non-ambiguous questions.}
    \label{fig:andor-byans}
\end{figure}

\begin{figure}[H]
    \centering
    \includegraphics[height=6cm]{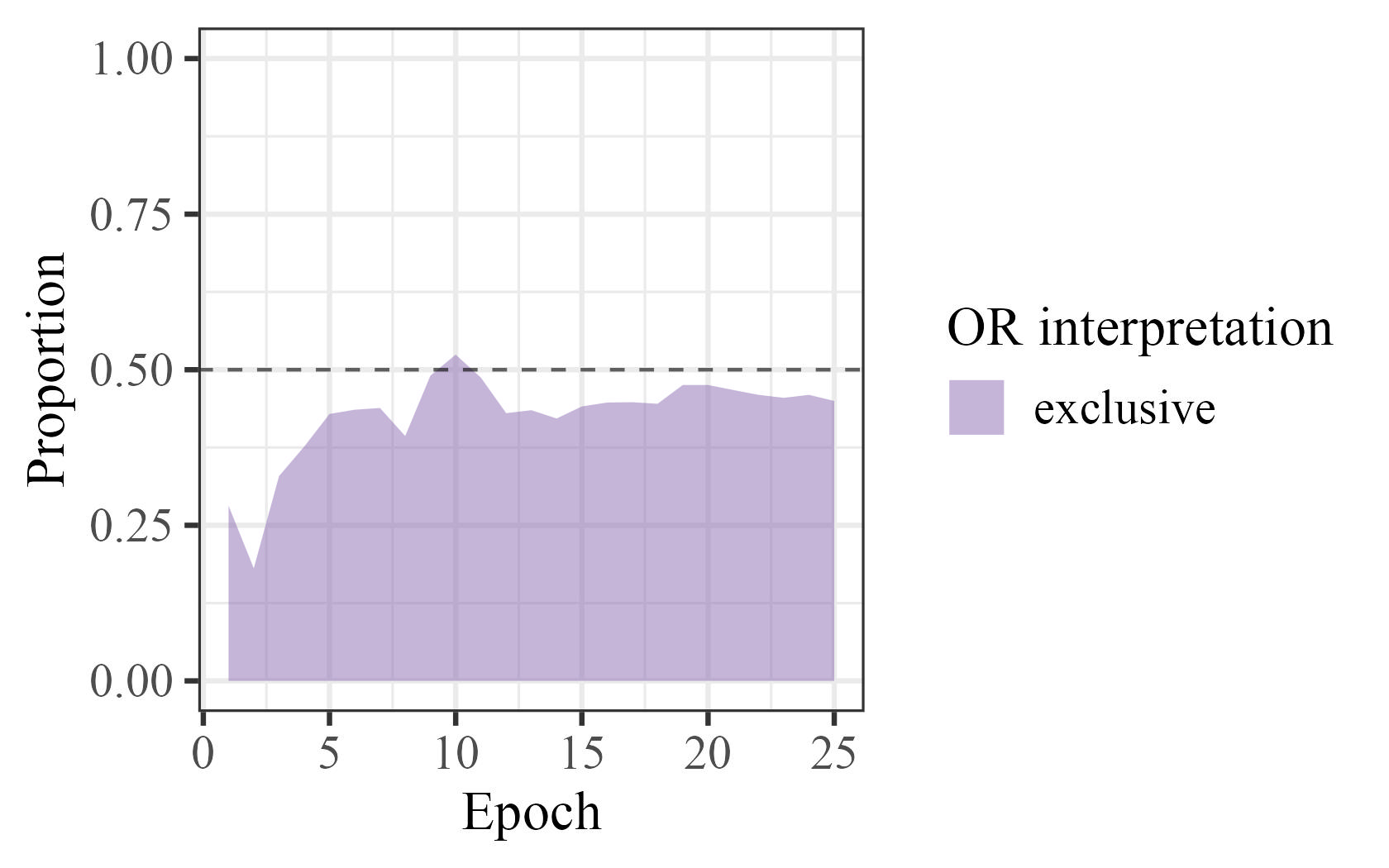}
    \caption[Experiment 1: proportion of exclusive (versus inclusive) OR interpretations.]{Experiment 1: Average proportion of exclusive (versus inclusive) interpretations of OR probe in ambiguous contexts, $(\alpha \land \beta)$. Overall standard deviation is  +/- 0.36, 3/5 runs learning to favor exclusive interpretation more than 50\% of the time.}
    \label{fig:or-exin}
\end{figure}

Next, Figure \ref{fig:andor-byans} shows the mean F1 score reported in the previous figure as a function of the answer type -- `yes' or `no' -- expected for each question for these probes. There is a clear asymmetry for both probes between questions in contexts requiring a `no' answer versus a `yes', and second, models performance in `yes' contexts then seems to drop after the second epoch. For AND, `yes' is expected in $(\alpha \land \beta)$ contexts and `no' otherwise. For OR, `yes' is expected in $(\alpha \land \neg \beta)$ and $(\neg \alpha \land \beta)$ contexts, while `no' is expected in $(\neg \alpha \land \neg \beta)$ contexts. Though models have no issue recognizing the answer in $(\neg \alpha \land \neg \beta)$, they struggle more when OR and AND expect opposing answers. This drop seems to also coincide with the rise of exclusive interpretations for OR in $(\alpha \land \beta)$ contexts as we see in Figure \ref{fig:or-exin}.

In Figure \ref{fig:or-exin}, we consider the proportion of inclusive versus exclusive interpretations of OR questions in the contexts where $(\alpha \land \beta)$ are both true. Importantly, the CLEVR dataset generative model hard-codes \textit{or} to be interpreted inclusively, in other words, all answers in the training data assume an inclusive \textit{or}. As we might expect, the models initially learn to favor inclusive interpretations. However, as learning progresses they start to interpret OR as exclusive more and more.

The differences in performance as a function of the answer types across AND - OR probes suggest that the models struggle more in contexts where AND questions and OR questions have conflicting answers, specifically, in $(\alpha \land \neg \beta)$ and $(\neg \alpha \land \beta)$ contexts. On the other hand, the results in contexts where $(\alpha \land \beta)$ are both true and both AND and OR should have the same answer (assuming an inclusive interpretation of \textit{or}), initially models seem to have no issues, but over time they start to favor exclusive interpretations for \textit{or} and struggle more with \textit{and} questions in the `yes' answer contexts. These results suggest that when determining the answer to a question containing \textit{and} or \textit{or}, models are also considering alternative questions that contain the other logical connective. In the cases where opposite answers for AND versus OR questions are expected, this attention to alternatives could lead to more uncertainty about the right answer. While in the case where the same answer is expected, it may instead be leading to a process akin to `reasoning about alternatives' where opposing logical operators should also have opposing answers, resulting in the rise of exclusive \textit{or}. We explore this hypothesis further in experiment 2, section \ref{sec:exp2}.

\begin{figure}[H]
    \centering
    \includegraphics[height=6cm]{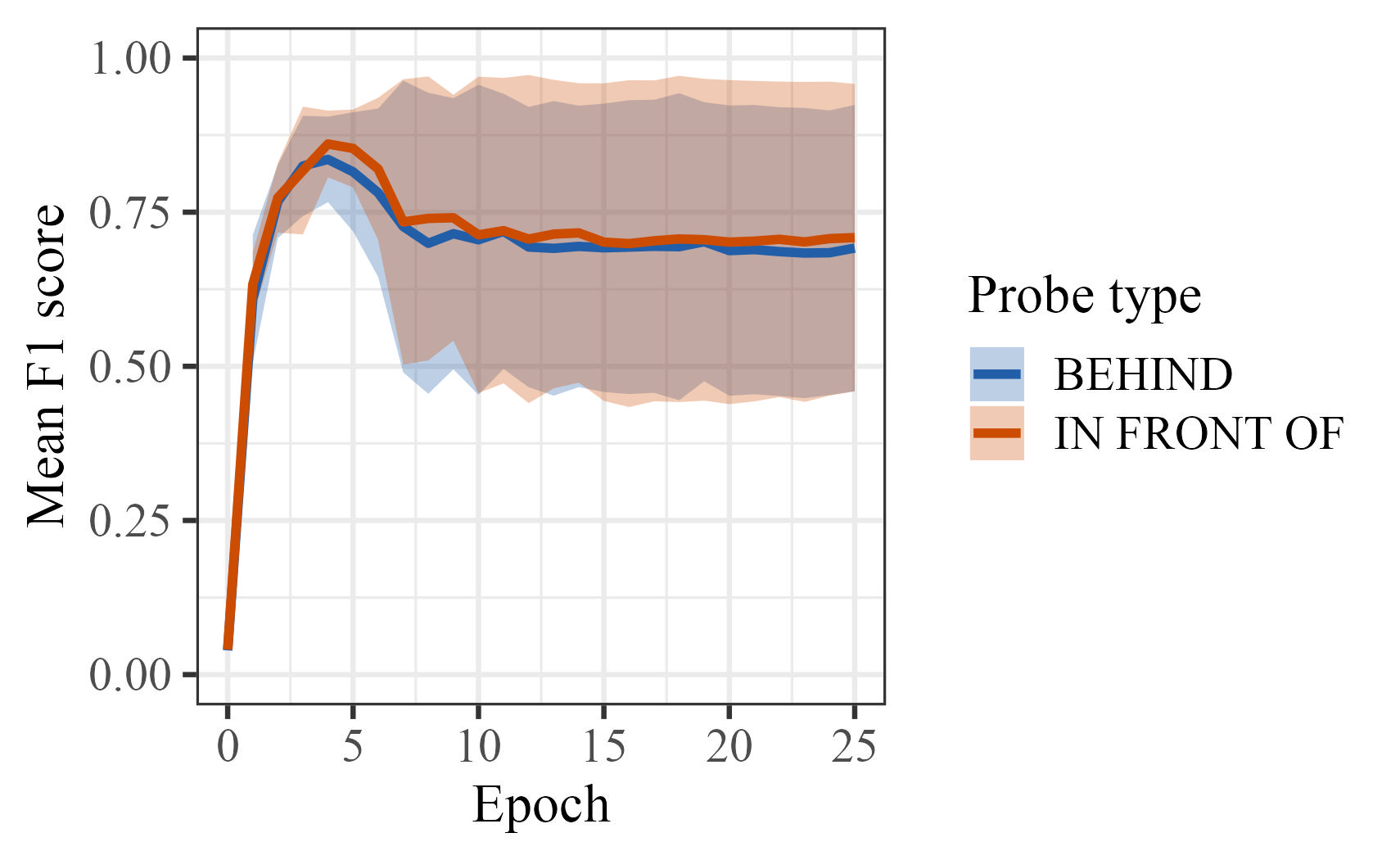}
    \caption[Experiment 1: F1 score on BEHIND - IN FRONT OF probes overall.]{Experiment 1: Mean F1 score on BEHIND - IN FRONT OF probes overall, shading represents standard deviation across 5 models.}
    \label{fig:behind-over}
\end{figure}

\paragraph{BEHIND - IN FRONT OF} Probe questions are all of form `Is the $X$ behind/in front of the $Y$?', and expect opposing answers as a function of the relative position of $X$ to $Y$. Figure \ref{fig:behind-over} shows the overall F1 scores of the models on both probes. There is more variation across random seed runs, though both BEHIND and IN FRONT OF seem to be learnt equally well within runs and performance is generally above chance.\footnote{We note that there is a slight drop in mean performance at the 6 epoch mark. Two of the five random seed runs seem to be causing this drop, while the other three continue increasing. In run 0, the model's performance on both BEHIND and IN FRONT OF drops specifically in the context of questions requiring `yes' answers, while in run 4, the opposite is true, dropping in the context of `no' answers. We do not know why this might be happening in these specific runs, but since most runs do not seem to have this problem, it may be safe to assume that these drops are due to the randomness introduced by different model initializations.} Unlike for AND and OR, Table \ref{tab:clevr-ans-freq} shows us that \textit{behind} and \textit{in front of} are used in a similar number of questions and expect `yes/no' answers at equal frequencies; we can therefore fairly compare models' relative performance on these words.

As with the previous probes, we also consider models' performance as a function of the answer type. Whether the context required a `yes' or `no' answer did not seem to matter for these probes as much as it did for others; models performed just as well in either context overall.

\begin{figure}[H]
    \centering
    \includegraphics[height=6cm]{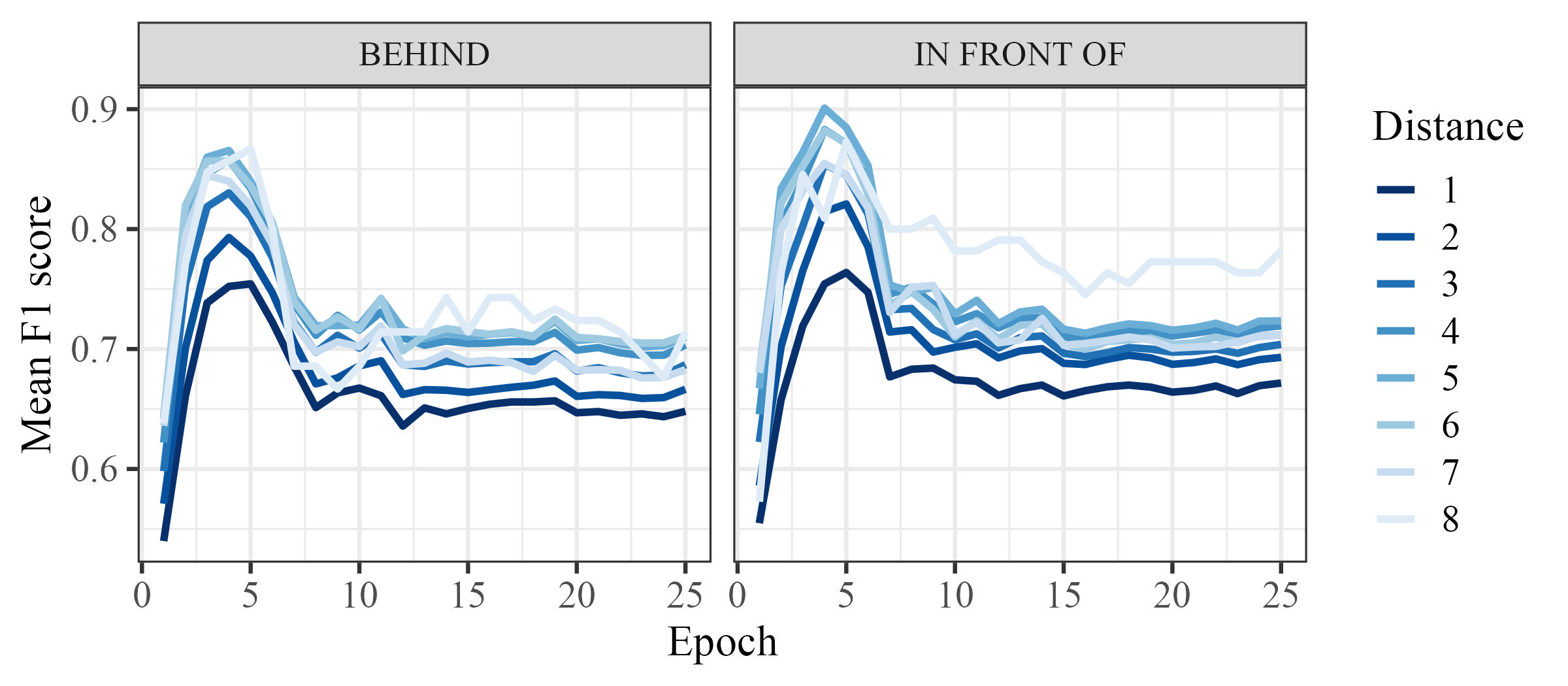}
    \caption[Experiment 1: F1 score on BEHIND - IN FRONT OF probes by distance.]{Experiment 1: Mean F1 score on BEHIND - IN FRONT OF probes as a function of the Euclidean distance between referents.}
    \label{fig:behind-bydist}
\end{figure}

In Figure \ref{fig:behind-bydist}, we look at how well models predict the correct answer to our BEHIND - IN FRONT OF probe questions as a function of the Euclidean distance between $X$ and $Y$ referents. The distances were calculated based on the coordinates of the center of each object provided in the metadata of each image. We then rounded the distances to the closest integer to bin our data into distance levels. Objects that have a Euclidean distance of 1 are so close that we expect one to partially occlude the other, while distances of 8 are as far apart as objects can be within a CLEVR image. As we can see from the figure, there is a very clear gradience in performance based on the distance between $X$ and $Y$, such that the further apart two objects are, the easier it is for the model to correctly interpret \textit{behind} and \textit{in front of}.

These results suggest the models can learn meaningful representations  \textit{behind} and \textit{in front of} such that they can interpret them in novel contexts. Furthermore, when these prepositions are equally frequent in models' input, they are learnt at the same rate. Importantly, models seem to learn a gradient semantic representation for the words as a function of the distance between referents, rather than the strict threshold based meaning which the CLEVR generative model uses.

\begin{figure}[H]
    \centering
    \includegraphics[height=6cm]{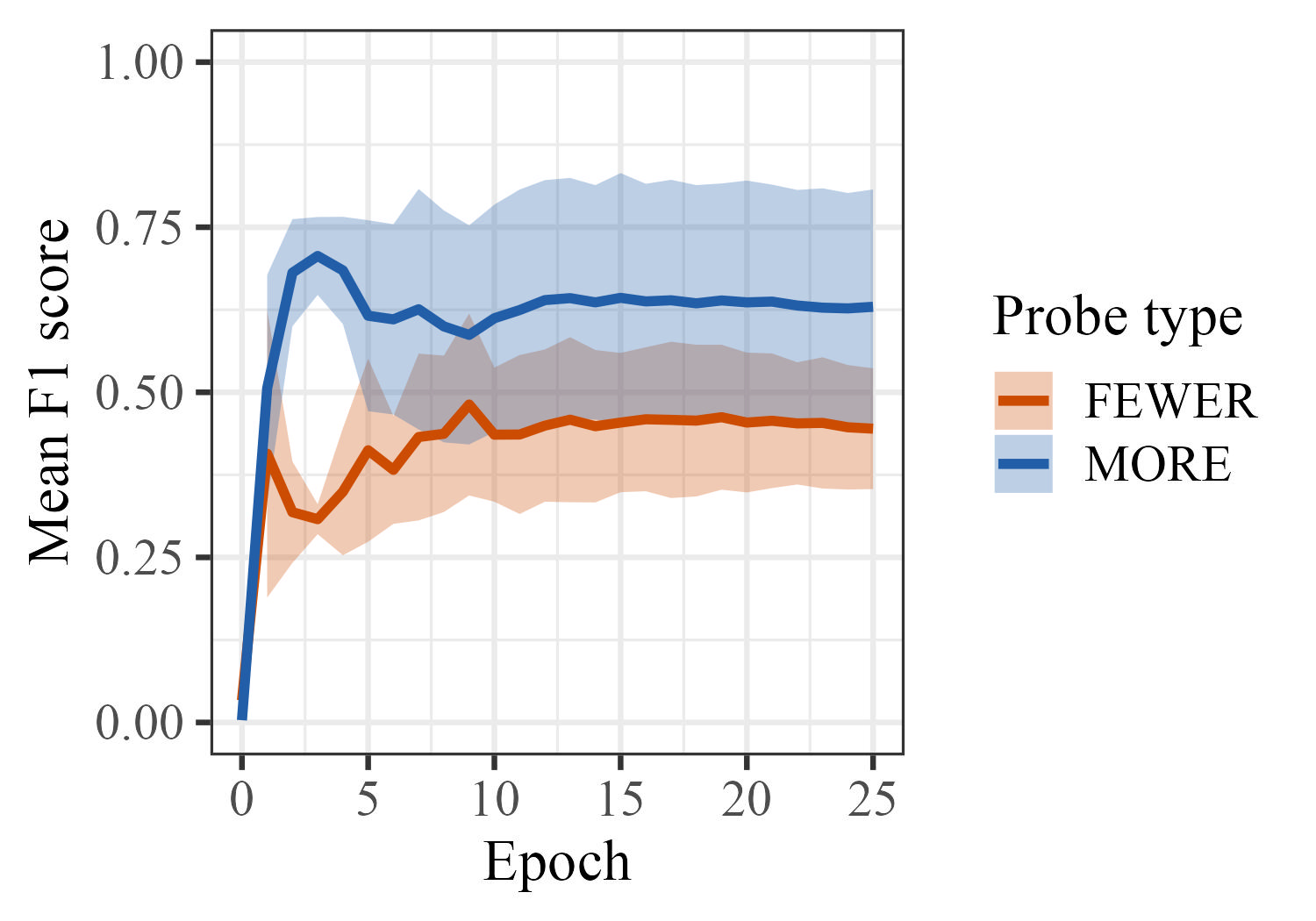}
    \caption[Experiment 1: F1 score on MORE - FEWER probes overall.]{Experiment 1: Mean F1 score on MORE - FEWER probes overall, shading represents standard deviation across 5 models.}
    \label{fig:more-over}
\end{figure}

\paragraph{MORE - FEWER} Probes are composed of questions of the form `Are there more/fewer of the $X$s than the $Y$s?'. For this analysis, we consider three contexts: when $|X| > |Y|$, $|X| < |Y|$, and $|X| = |Y|$. In the first two contexts, MORE and FEWER questions expect opposite answers, while in the third context where there is no difference in the number of $X$s and $Y$s, they expect the same answer, `no'. Figure \ref{fig:more-over} presents the overall F1 scores of models on both probes. This initial plot suggests that MORE is learnt first and may be overall easier than FEWER.

\begin{figure}[H]
    \centering
    \includegraphics[height=6cm]{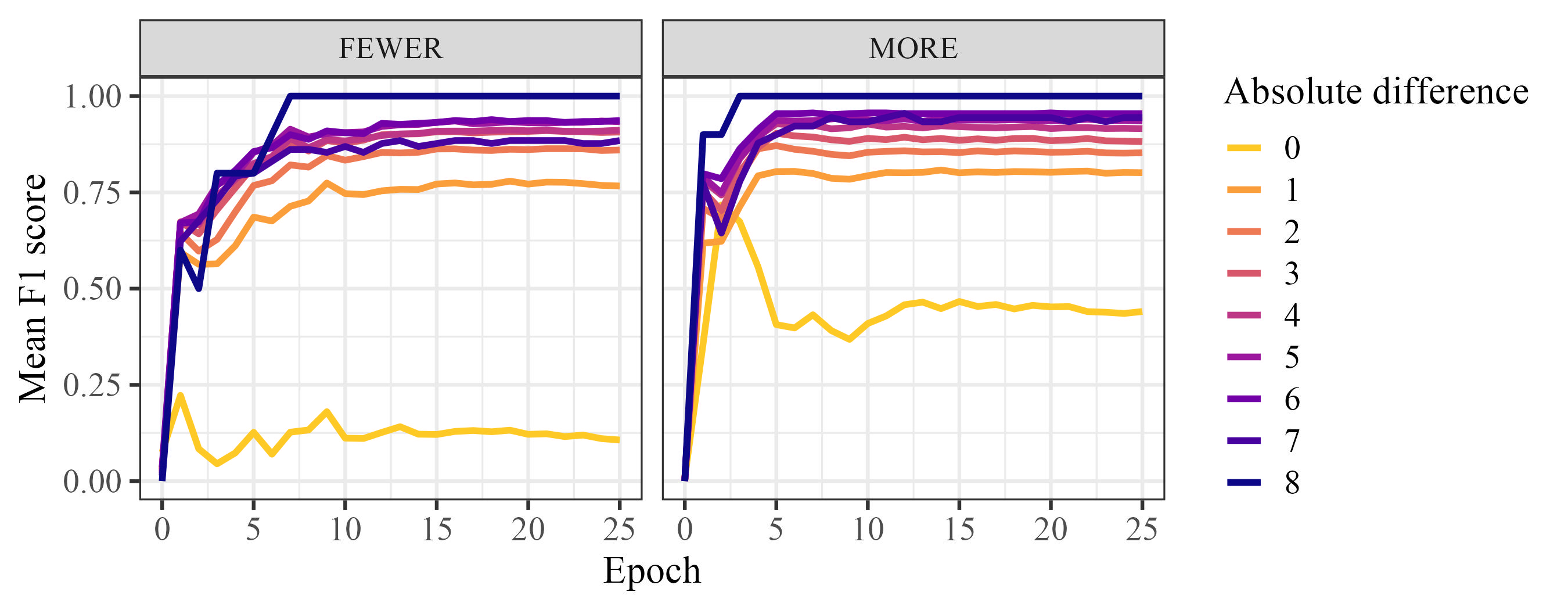}
    \caption[Experiment 1: F1 score on MORE - FEWER probes by absolute count difference.]{Experiment 1: Mean F1 score on MORE - FEWER probes by absolute difference in the number of objects in each referent class.}
    \label{fig:more-bynum}
\end{figure}

Next, we plot accuracy on probes as a function of the absolute difference between the number of $X$s and $Y$s, $absolute(|X| - |Y|)$, Figure \ref{fig:more-bynum}. Models clearly struggle with both MORE and FEWER questions specifically when the difference is 0, or $|X| = |Y|$, performing below chance in this context. In all other cases, whether the answer is `yes' or `no', models correctly answer questions over 75\% of the time. Yet again, performance for these probes is gradient. Models correctly interpret both \textit{more} and \textit{fewer} more often as a function of the difference in number between the two referent classes. The larger the difference, the easier it is for the model to correctly judge whether there are \textit{more} or \textit{fewer} of a given class of referents. Additionally, models poorer performance on FEWER probe questions overall seen in the previous plot seems to be isolated to the contexts where $|X| = |Y|$.

\begin{figure}[H]
    \centering
    \includegraphics[height=6cm]{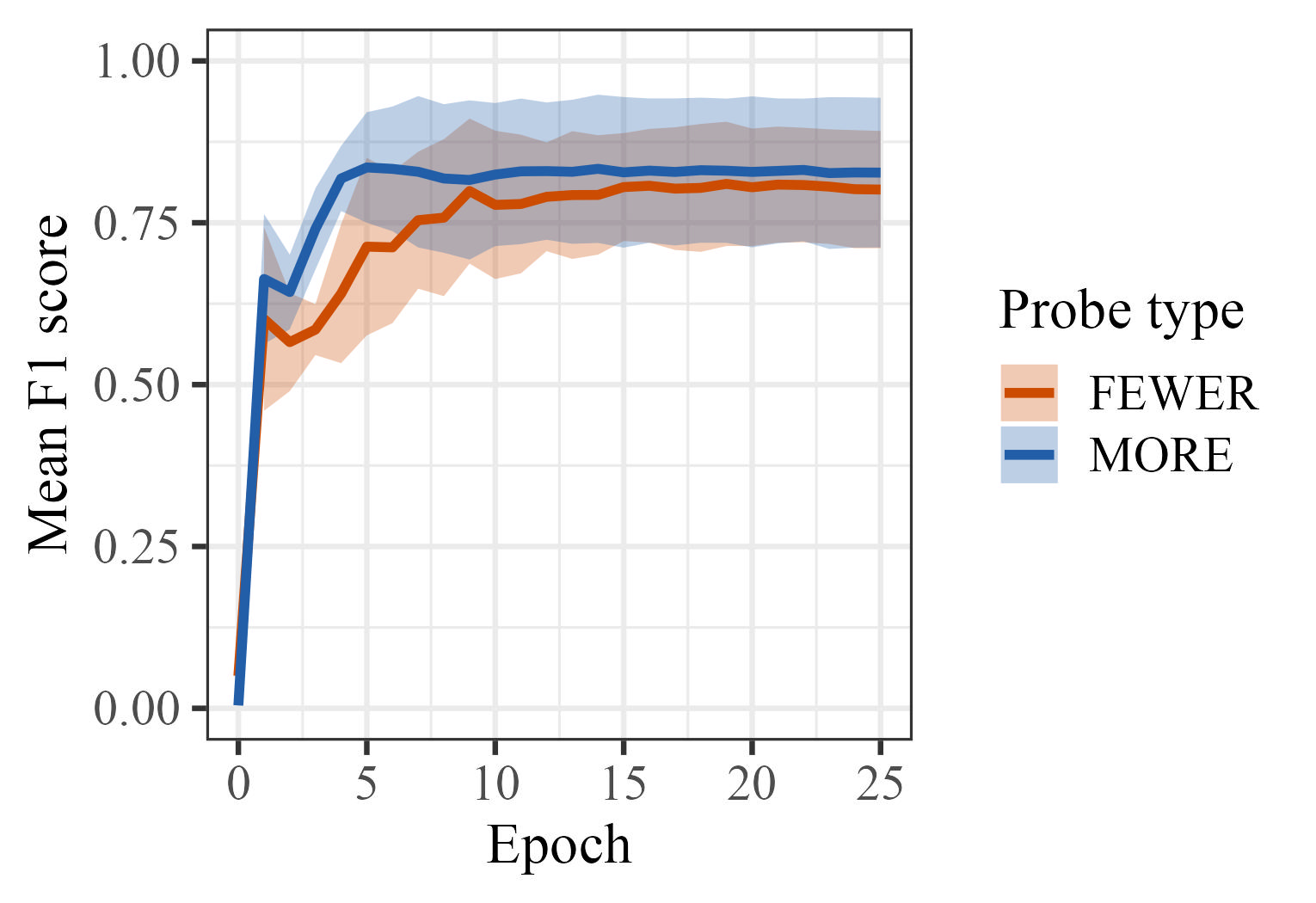}
    \caption[Experiment 1: F1 score on MORE - FEWER probes overall without even cases.]{Experiment 1: Mean F1 score on MORE - FEWER probes overall excluding context where $|X| = |Y|$, shading represents standard deviation across 5 models.}
    \label{fig:more-overno0}
\end{figure}

In fact, if we remove all probe questions where $|X| = |Y|$ and consider the overall  of models again in Figure \ref{fig:more-overno0}, we see a very different picture than our original Figure \ref{fig:more-over}. Models have almost equally high performance on both probes,  still learning \textit{more} slightly earlier than \textit{fewer}.

These results suggest that models learn reasonable meaning representations for both \textit{more} and \textit{fewer} and furthermore, that these representations are gradient as a function of the difference in number between two referent categories, rather than being based on strict thresholds, which the CLEVR generative model uses. However, models struggled in contexts where $|X| = |Y|$ specifically. We hypothesise that this may be because they are exposed to a third alternative numerical reasoning expression during training, \textit{equal/same number}. This alternative expression expects an opposing answer in these contexts. Like with AND and OR, models may be considering the existence of alternative propositions when trying to answer these questions, leading to more uncertainty in the context where the difference in number between $X$s and $Y$s is the smallest. We explore this hypothesis further in the next experiment.

\subsection{Interim conclusion}

Our first experiment examined the first research question: How do visually grounded question answering models learn to represent and interpret function words and do the representations they learn generalize to unseen linguistic and visual contexts? We found
that models learnt gradient interpretations for function words requiring spatial and numerical, \textit{behind}, \textit{in front of}, \textit{more}, and \textit{fewer}. Additionally, we found early evidence that models consider alternative logical connectives when determining the meaning of expressions containing \textit{and} and \textit{or}. This behaviour may be leading models to interpret \textit{or} as exclusive in an increasing number of context. Further experimentation is necessary to test this hypothesis.

\section{Experiment 2: The effect of alternatives on reasoning}\label{sec:exp2}

Does the existence of alternative expressions in each reasoning pair affect their acquisition or are the meanings of function words acquired in isolation? Following our results from the previous experiment, we hypothesized that models could be considering alternative questions and answers which use opposing or parallel function words when they compute the probability of the answer to a given question. This process akin to `reasoning about alternatives' could then explain the performance patterns we observed specifically for the AND - OR probes, as well as the MORE - FEWER probes. Unlike BEHIND - IN FRONT OF which always expect opposing answers, AND - OR and MORE - FEWER pairs both have contexts where they expect the same answer and others where they do not.

The existence of alternative expressions may lead to uncertainty in model predictions in one of two ways. First, if models observe that \textit{and} and \textit{or} are interpreted the same in a set of contexts, then they may begin to expect them to also mean something similar in contexts where they actually should have opposing answers. Second, if models instead observe that they have opposing answers in a set of contexts, then they may instead begin to expect them to mean something different also in contexts where in fact they should be interpreted the same way. In either case, the existence of the alternative expression (\textit{and} in the case of  \textit{or}, and \textit{or} in the case of \textit{and}) is what leads models to answer incorrectly, showing evidence of this process. Our second experiment tests our theory and answer our second research question.

\subsection{Setup}

As in experiment 1, we train five MAC models initialized using different random seeds. Unlike the previous experiment, however, we manipulate the training data to remove alternative function words which we believe affected the probe performance for OR, AND, MORE, and FEWER. Specifically, we remove all questions from the training data which contain the word \textit{and} and then evaluate model performance on the OR probe. We repeat this process and create a version of CLEVR where we remove all instances of \textit{or} and then evaluate models on the AND probe. Finally, we create a version without \textit{equal/same number of} and evaluate the models on MORE and LESS probes. By removing \textit{and}, we want to see if the model will correctly learn the semantics of \textit{or} and favor inclusive interpretations when the alternative logical connective is not present. By removing \textit{or}, we want to make sure models learn to correctly interpret \textit{and} regardless of the answer context. Finally, by removing \textit{equal/same number of} and its derivatives, we would like to see if the models can correctly learn to use \textit{more} and \textit{fewer} in contexts where $|X| = |Y|$, when the alternative proposition that there are an \textit{equal} amount of them is no longer available. For each of these different subsampled training datasets and evaluation probes, we train models for 25 epochs and evaluate performance on probes at each epoch.

\subsection{Results}
Since the results for the OR probe and AND probe come from different models trained on different subsampled datasets, we report their results separately for this experiment.

\begin{figure}[H]
    \centering
    \includegraphics[height=6cm]{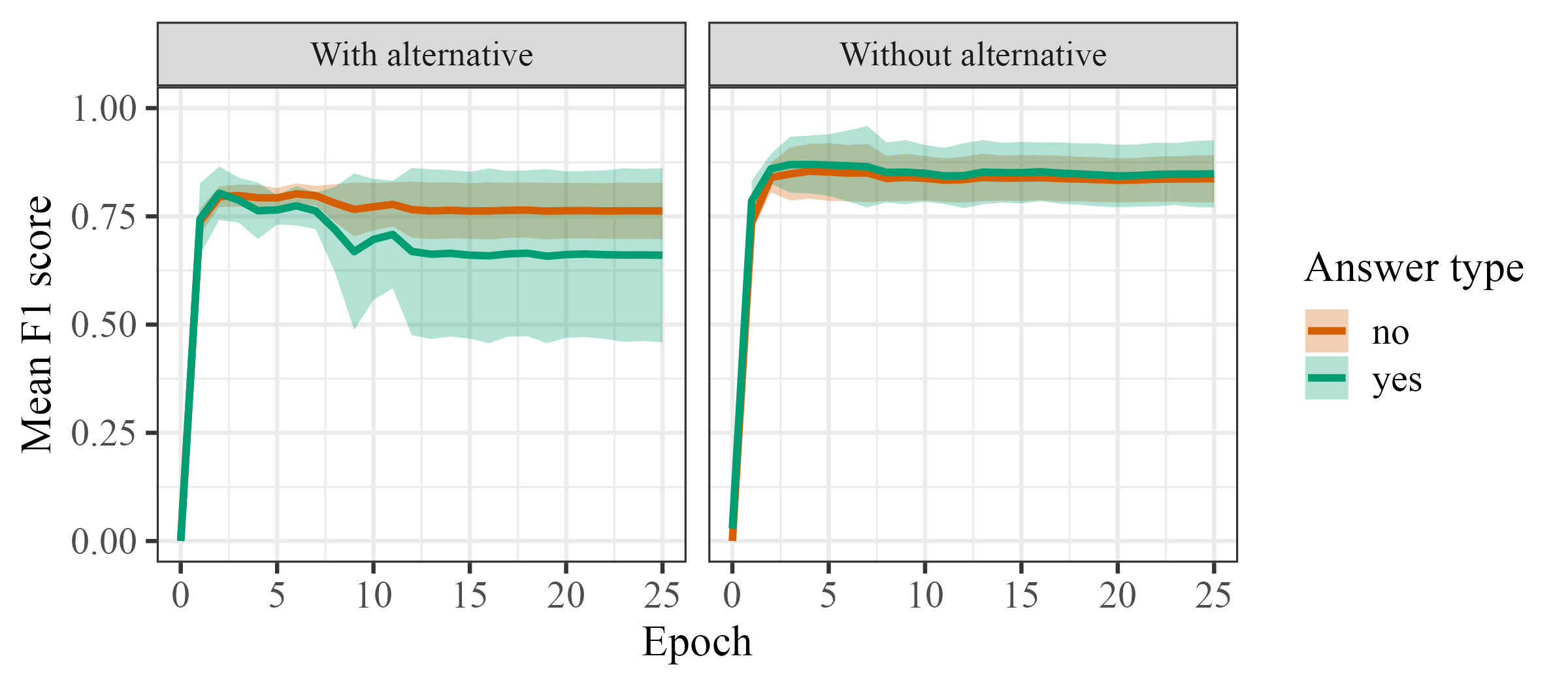}
    \caption[Experiment 2: Accuracy on OR probe by answer type.]{Experiment 2: Mean F1 score on OR probe by answer type in non-ambiguous questions when trained on data with the alternative expression \textit{and} from experiment 1 versus without this alternative in experiment 2.}
    \label{fig:or-byans-exp2}
\end{figure}

\paragraph{OR} Models reach a higher overall mean F1 score on unambiguous OR probe questions when trained on data without \textit{and}. Comparing model performance when trained with and without \textit{and} as a function of the answer type expected in Figure \ref{fig:or-byans-exp2}, it is clear that when we remove the alternative expression models no longer struggle in contexts expecting a `yes' answer as they did in experiment 1.

\begin{figure}[H]
    \centering
    \includegraphics[height=6cm]{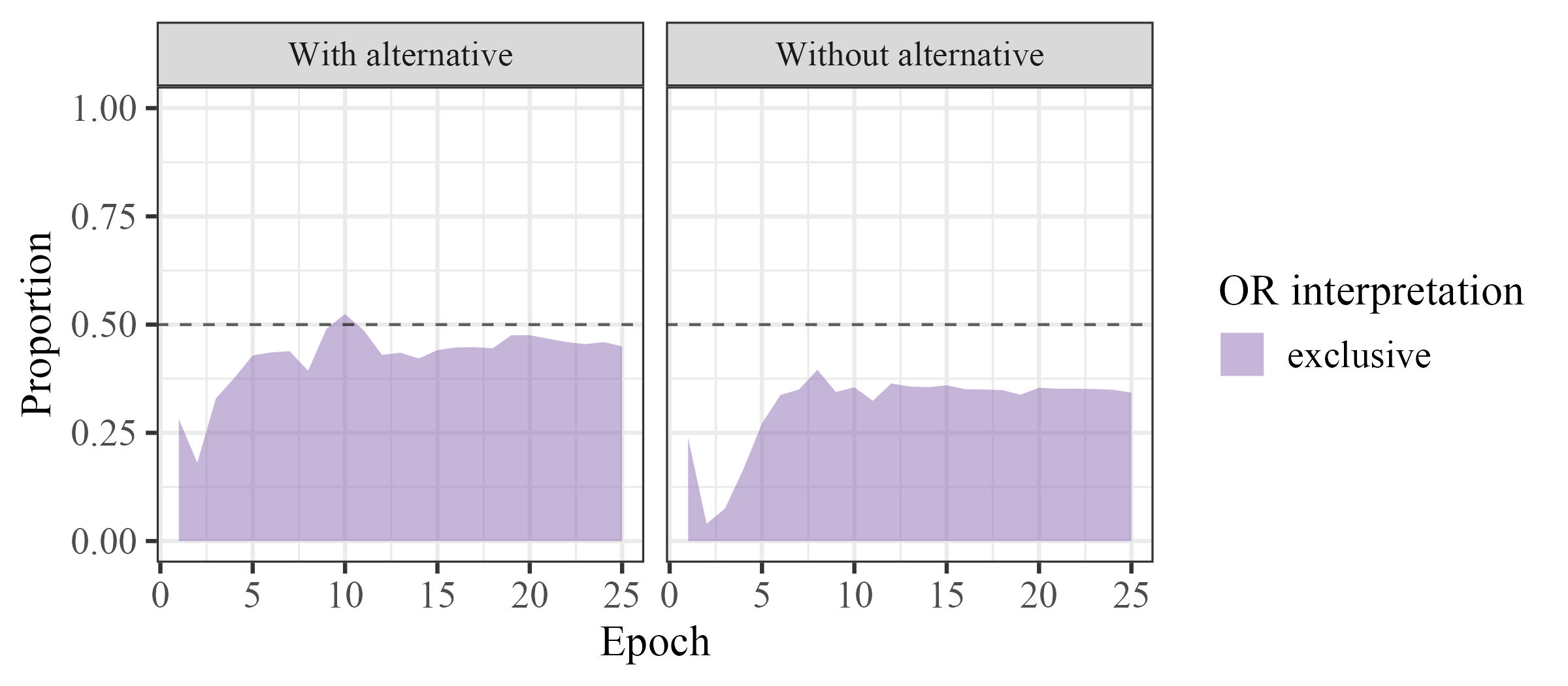}
    \caption[Experiment 2: proportion of exclusive (versus inclusive) OR interpretations.]{Experiment 2: Proportion of exclusive (versus inclusive) interpretations of OR probe in ambiguous contexts, $(\alpha \land \beta)$, when trained on data with the alternative expression \textit{and} from experiment 1 versus without this alternative in experiment 2. Overall standard deviation for experiment 2 is +/- 0.3, 4/5 runs favoring inclusive interpretations.}
    \label{fig:or-exin-exp2}
\end{figure}

As for probe questions containing \textit{or} in ambiguous contexts where inclusive-or and exclusive-or interpretations predict opposing answers, we no longer see as strong of a progressive rise in exclusive interpretations, instead settling on average with around 70\% of ambiguous questions being answered with inclusive `yes' answers, Figure \ref{fig:or-exin-exp2}.

When the alternative logical connective \textit{and} is not present, models have no difficulty learning the semantics of \textit{or}. Since the CLEVR generative model defines \textit{or} as inclusive, when no pragmatic alternative is present, models also learn to interpret \textit{or} inclusively. These results support the hypothesis that the rise in exclusive interpretations seen in experiment 1 is due to some form of competition between \textit{or} and the available alternative \textit{and}.

\begin{figure}[H]
    \centering
    \includegraphics[height=6cm]{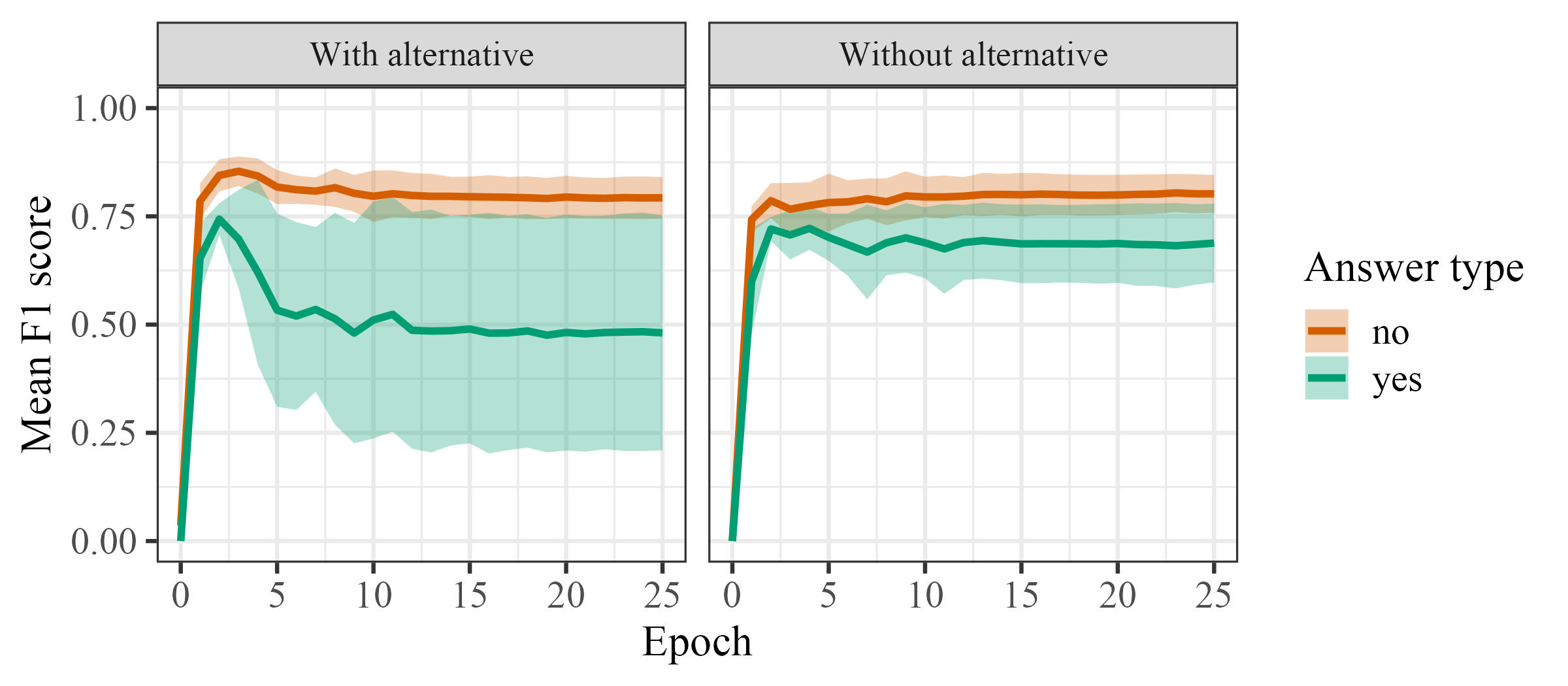}
    \caption[Experiment 2: F1 score on AND probe by answer type.]{Experiment 2: Mean F1 score on AND probe by answer type when trained on data with the alternative expression \textit{or} from experiment 1 versus without this alternative in experiment 2.}
    \label{fig:and-byans-exp2}
\end{figure}

\paragraph{AND} Probe results come from models trained on a subsampled version of CLEVR where all instances of \textit{or} have been removed. Models had better F1 scores on AND probe questions when the alternative logical connective was removed than when both were present in experiment 1. In the absence of \textit{or}, models learned to correctly interpret \textit{and} regardless of the truth value context, Figure \ref{fig:and-byans-exp2}.

Models can learn the meaning of the logical connective \textit{and} correctly and then generalize it to interpret this word in novel contexts. If the alternative logical connective for disjunction is present, like in experiment 1, then the models may struggle more, as they seem to consider the existence of this alternative when trying to determine the intended meaning of \textit{and}. This difficulty disappears if the alternative is no longer present.

\begin{figure}[H]
    \centering
    \includegraphics[height=6cm]{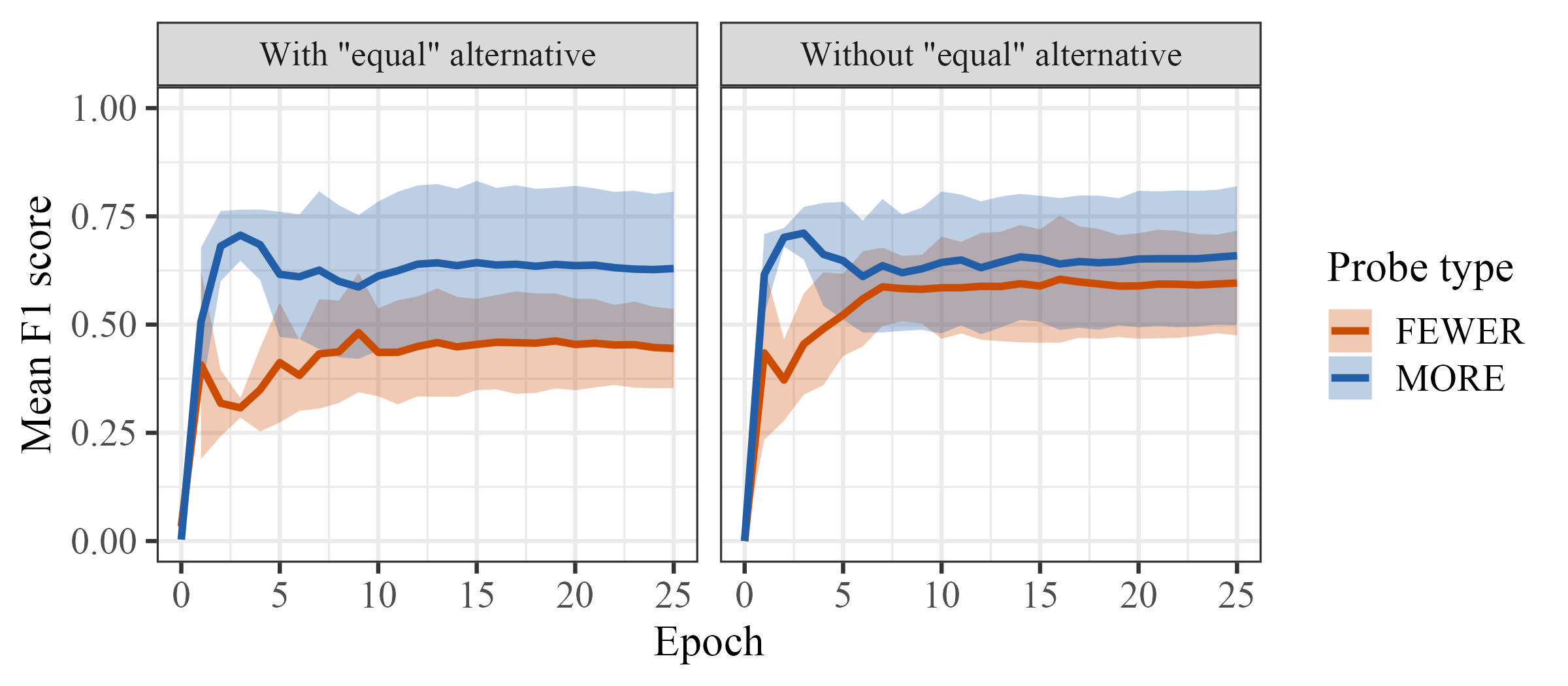}
    \caption[Experiment 2: F1 score on MORE - FEWER probes overall.]{Experiment 2: Mean F1 score on MORE - FEWER probes overall when trained on data with the alternative expression \textit{equal} from experiment 1 versus without this alternative in experiment 2. Shading represents standard deviation across 5 models.}
    \label{fig:more-over-exp2}
\end{figure}

\paragraph{MORE - FEWER} Probe results from experiment 1 showed that models struggled to correctly interpret \textit{more} and \textit{fewer} in the context where there were an equal number of the two referent categories being compared. We hypothesized that models may have struggled in this context because there existed alternative questions that asked whether there were an \textit{equal} number of $X$s and $Y$s in the training data. To test this hypothesis, we trained models on a subsampled version of CLEVR where we removed all questions that asked about number equality. Figure \ref{fig:more-over-exp2} shows the overall performance of these models on both probes when trained with and without this alternative \textit{equal} expression. F1 scores on FEWER questions have definitely risen in comparison to experiment 1, though results for MORE look quite similar.

\begin{figure}[H]
    \centering
    \includegraphics[height=6cm]{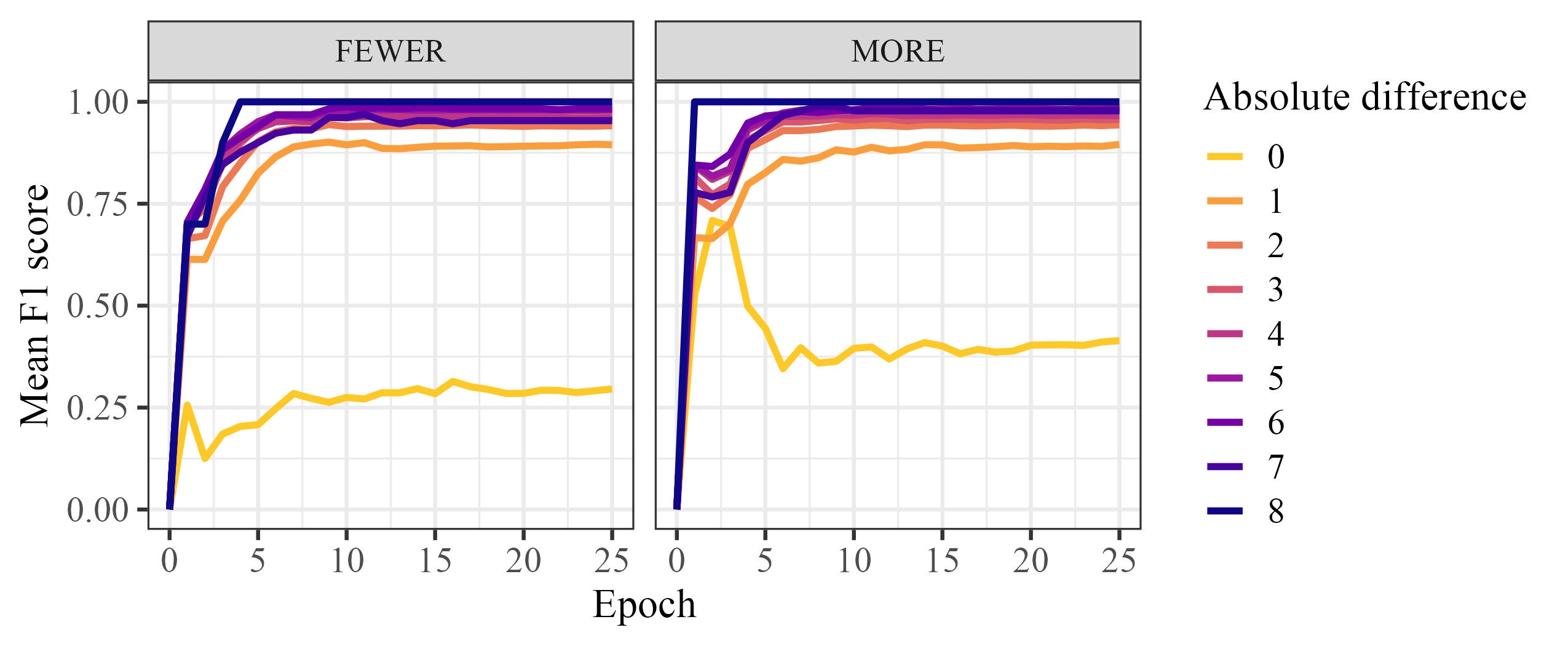}
    \caption[Experiment 2: F1 score accuracy on MORE - FEWER probes by absolute count difference.]{Experiment 2: Mean F1 score on MORE - FEWER probes by absolute difference in the number of objects in each referent class when trained without the alternative \textit{equal} expression.}
    \label{fig:more-bynum-exp2}
\end{figure}

However, when we consider model performance on questions as a function of the absolute difference in number between the compared referent categories in Figure \ref{fig:more-bynum-exp2}, models still struggle in contexts where $|X| = |Y|$. They do better overall in all other contexts.

\begin{figure}[ht]
    \centering
    \includegraphics[width=\textwidth]{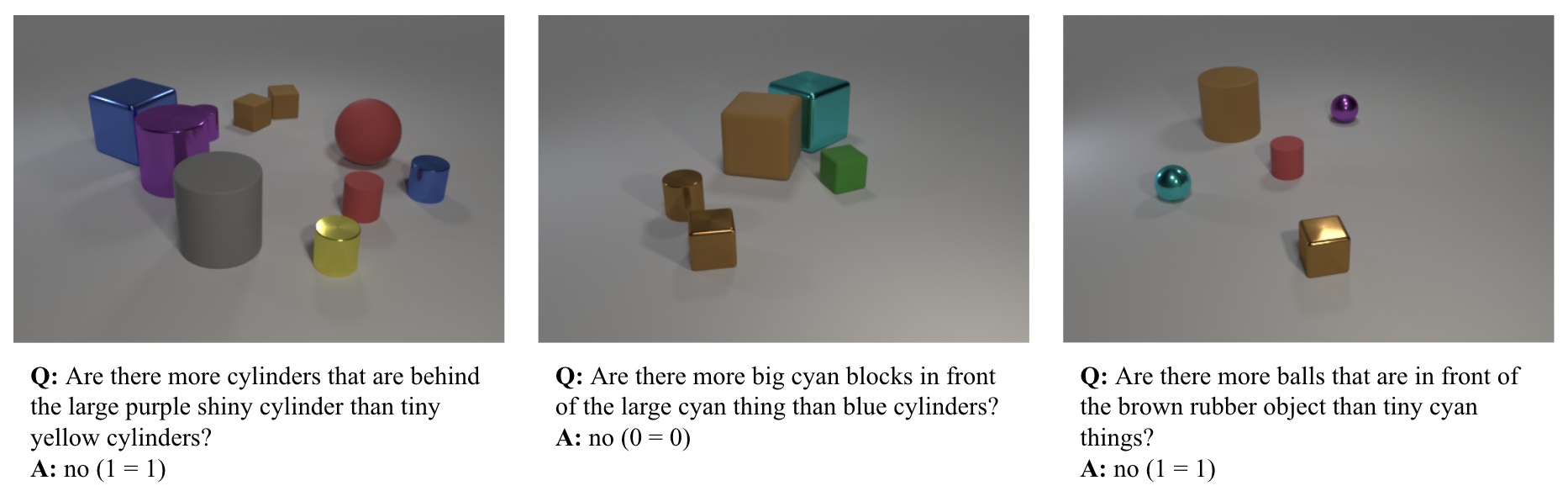}
    \caption[Example CLEVR training questions with \textit{more} in equality contexts.]{Example CLEVR training questions with the word \textit{more} in the context where $|X| = |Y|$}.
    \label{fig:example-more0}
\end{figure}

Unlike with AND and OR, removing the pragmatic alternative did not solve our issue with FEWER and MORE. After carefully scrutinising the training data from CLEVR, it became apparent that \textit{more/fewer} rarely appeared in contexts where $|X| = |Y|$ and only when they were part of more complex question templates. Figure \ref{fig:example-more0} shows example questions with \textit{more} in the context where $|X| = |Y|$ taken from the CLEVR train data. Thus, the issues we see with probe performance in this context may simply be due to our choice of template and the idiosyncrasies in the distribution of \textit{more} and \textit{fewer} in the CLEVR training data.

\subsection{Interim conclusion}

This experiment examined our second research question: does the existence alternative expressions in each reasoning pair affect their acquisition or are the meanings of function words acquired in isolation? We found that in the absence of a logical alternative, models correctly learnt to generalize the meaning of conjunction and disjunction. Our findings confirm our hypothesis that the presence, or absence, of a pragmatic alternative can affect how models learn to interpret logical connectives \textit{and} and \textit{or}. Next, we will evaluate how the frequency of different function words may also affect how models learn their meanings.

\section{Experiment 3: The effect of frequency on learning}

Our third and final experiment considers the effect of word frequency on the order in which function words are learnt. We address our third research question: does the order in which function words are acquired by models resemble that of children -- and are some of these ordering effects simply the result of frequency in the input or are there other conceptual factors at play?

\subsection{Setup}

We again trained five MAC models initialized using different random seeds for a total of 25 epochs and consider their performance on semantic probes throughout training. Our main manipulation that differentiates this experiment from the others is the training data. As in experiment 2, we use a subsampled version of the CLEVR training questions. This time, we created a version of CLEVR where the relative frequencies of the target function words matched their relative frequencies across all English child-directed utterances from the CHILDES repository \cite{macwhinney2000CHILDES}.

The CHILDES repository is a collection of open-source transcripts, recordings, and videos of child-caregiver/experimenter interactions from a wide range of studies dating as far back as the 1950s. Children in these studies vary in age between 9 months and 5 years old, the median being about 3 years. Using the childes-db API \cite{sanchez2018childes} to access the data, we isolated all of the English transcript corpora available. We then filtered each to isolate all utterances that were not said by the child, representing a sample of the linguistic input the child was exposed to. We used this corpus to calculate the relative frequencies in children's input of the function words we are interested in. The corpus contained a total of 16,062,386 word tokens.

We considered the relative frequencies of our function words within each contrasting pair rather than their relative frequencies overall as it would not have been possible to extract a reasonably sized subsampled version of the CLEVR training data otherwise. One of the main difficulties we ran into when trying to subsample from the CLEVR dataset was that these function words often appeared in overlapping sets of questions, so changing the frequency of one word by subsampling questions would inadvertedly affect another's frequency. Nonetheless, we managed to create a version of the CLEVR training data that almost reproduced the relative frequencies of the CHILDES data and was of a reasonable size, containing 545,681 training questions (9,652,086 tokens). Table \ref{tab:clevr-childes-freq} shows the exact word counts and frequencies of both the CHILDES and subsampled CLEVR training datasets.

\begin{table}
    \centering
    \begin{tabular}{|l|c|c||c|c|}
    \hline
    & \multicolumn{2}{c ||}{CHILDES} & \multicolumn{2}{c |}{CLEVR subsampled}\\
word pair & raw counts & frequency  & raw counts & frequency\\
         \hline
         and & 217,497 & 90.45\% & 81,506 & 90.45\% \\
         or & 22,975 & 9.55\% & 8,610 & 9.55\% \\
         \hline
         behind & 2,954 & 79.62\% & 113,881 &  74.36\% \\
         in front of & 756 & 20.38\% & 39,260 & 25.64\% \\
         \hline
         more & 23,406 & 99.10\% & 11,570 & 99.10\% \\
         fewer/less & 212 & 0.90\% & 105 &  0.90\% \\
         \hline
    \end{tabular}
    \caption[Word frequencies in experiment 3 training data.]{Relative frequencies of each function word pair in the CHILDES and subsampled CLEVR training data for experiment 3.}
    \label{tab:clevr-childes-freq}
\end{table}

\subsection{Results}

\begin{figure}[H]
    \centering
    \includegraphics[height=6cm]{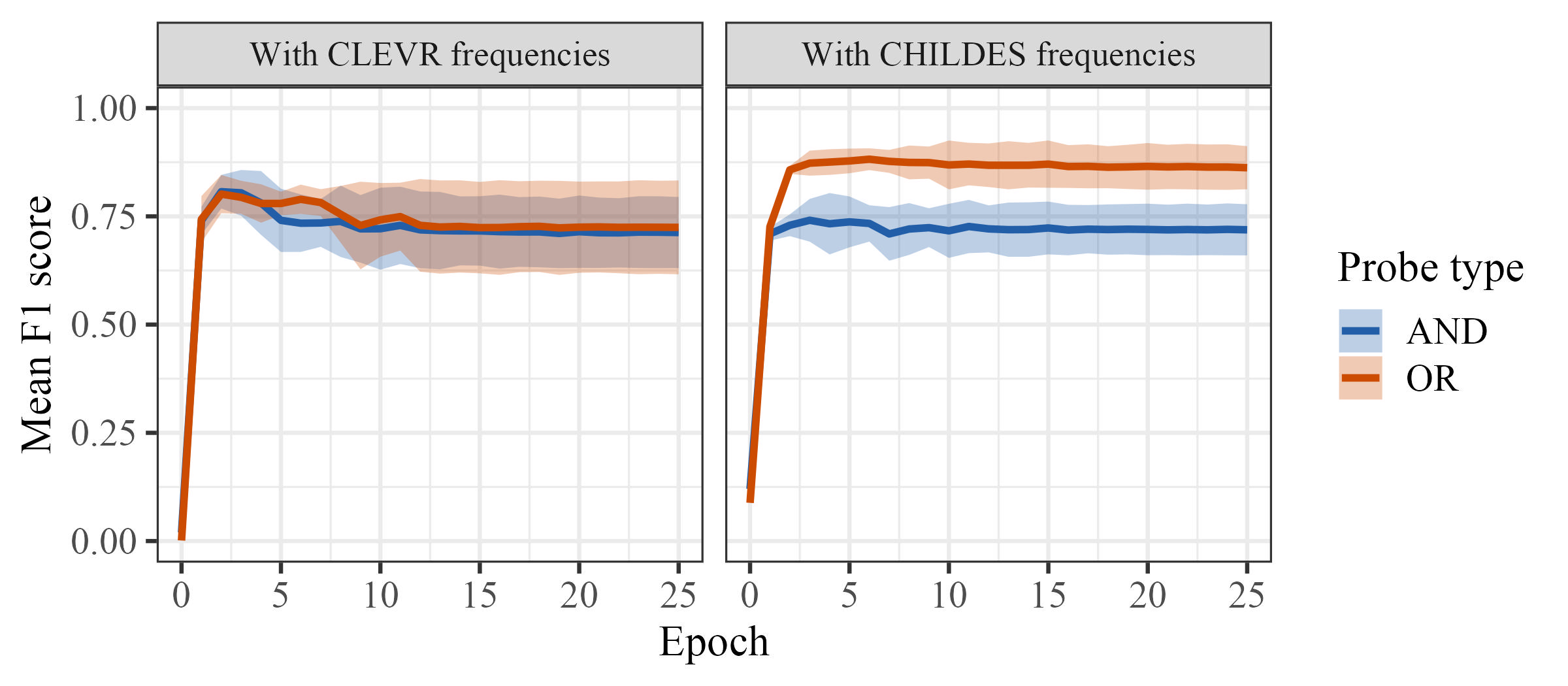}
    \caption[Experiment 3: F1 score on AND - OR probes overall.]{Experiment 3: Mean F1 score on AND - OR probes overall in non-ambiguous questions when trained on the original CLEVR dataset and the subsampled version with CHILDES-like frequencies. Shading represents standard deviation across 5 models.}
    \label{fig:andor-over-exp3}
\end{figure}

\paragraph{AND - OR} Figure \ref{fig:andor-over-exp3} compares the overall performance of the models on each of these probes in non-ambiguous questions (i.e. excluding OR questions in $\alpha \land \beta$ contexts) in experiment 1 with the original CLEVR dataset frequencies and the current experiment with CHILDES-like relative frequencies. These words have a very uneven distribution in the subsampled CLEVR version like CHILDES, \textit{and} being much more prominent than \textit{or} in children's input. Interestingly, even with this frequency imbalance, models seem to do quite well on both our AND and OR semantic probes, suggesting that even with a reduced number of training examples containing \textit{or}, they are still learning a reasonable representation for this word that allows them to generalize its meaning to unseen contexts.

\begin{figure}[H]
    \centering
    \includegraphics[height=6cm]{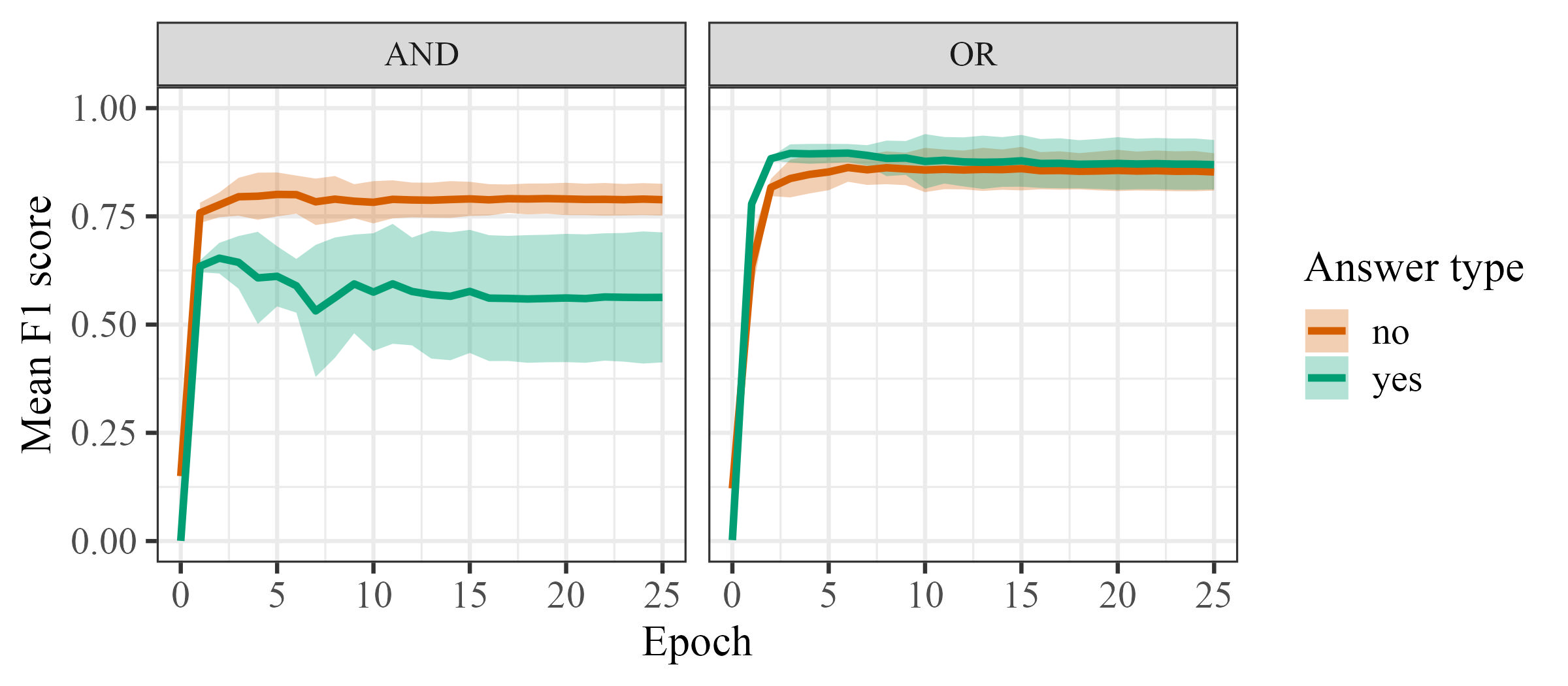}
    \caption[Experiment 3: F1 score on AND - OR probes by answer type.]{Experiment 3: Mean F1 score on AND - OR probes by answer type in non-ambiguous questions when trained of subsampled CLEVR.}
    \label{fig:andor-byans-exp3}
\end{figure}

This observation is confirmed when we consider the current models' mean F1 score by answer type, `yes' or `no', where OR probe performance is the same regardless of context, Figure \ref{fig:andor-byans-exp3}. As for the AND probe, models seem to be performing as it did in experiment 1, struggling more in contexts requiring `yes' as an answer.

\begin{figure}[H]
    \centering
    \includegraphics[height=6cm]{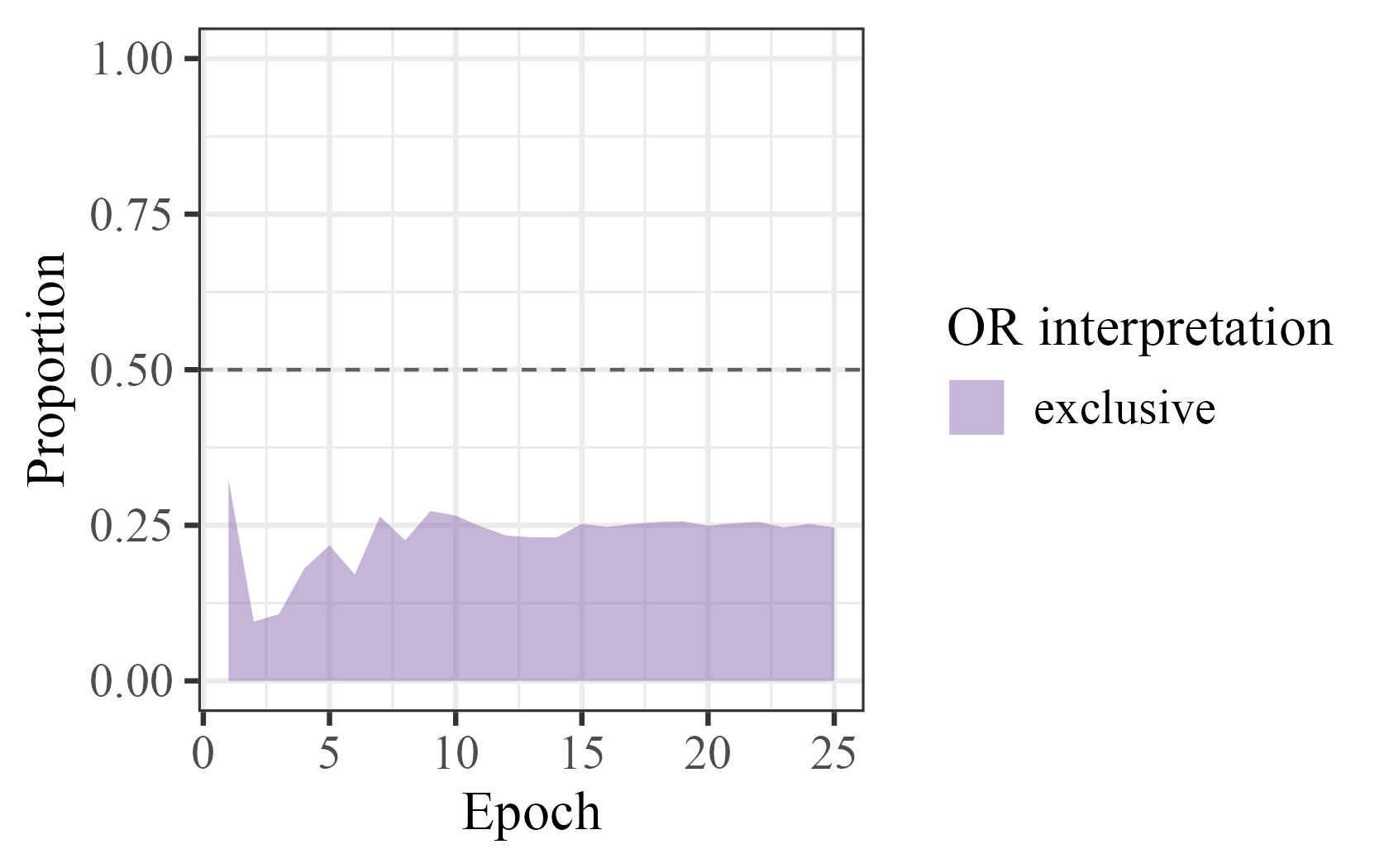}
    \caption[Experiment 3: proportion of exclusive versus inclusive OR interpretations.]{Experiment 3: Proportion of exclusive versus inclusive interpretations of OR probe in ambiguous contexts, $(\alpha \land \beta)$, when trained of subsampled CLEVR. Overall standard deviation is +/- 0.19, all runs favoring inclusive interpretations.}
    \label{fig:or-exin-exp3}
\end{figure}

In the case of ambiguous OR questions, in $\alpha \land \beta$ contexts, models clearly prefer inclusive answers; we see no rise in exclusive interpretations like the one seen in experiment 1, see Figure \ref{fig:or-exin-exp3}.

If performance on these probes were solely a function of the frequency of these words in models' input, we would expect their performance on the OR probe to decrease between experiment 1 and experiment 3, but as we saw in Figure \ref{fig:andor-over-exp3} this is not what happens. Furthermore, if the effect of being sensitive to possible alternative expressions was also proportional to the frequency of these alternatives, we might also expect to see a stronger effect of the alternative \textit{and} on OR probe results and an increase in inclusive interpretations for \textit{or}, but again, we do not see this effect. It seems to have been stronger in experiment 1 when \textit{and} and \textit{or} were about equally frequent. The more uniform distribution between these words in experiment 1 could have lead to more uncertainty overall. This explanation is further supported by the smaller standard deviations we see in Figure \ref{fig:andor-over-exp3} for models trained on the CHILDES-like frequencies versus those trained on the original CLEVR dataset. Another possible explanation that should not be discounted is that in downsampling questions containing \textit{or} in the training set, we may have simply reduced the diversity of contexts seen for \textit{or} in favor of contexts that resembled our probe template more, such that the models now had less uncertainty specifically about the meaning of \textit{or}.

As for AND, models' performance in experiment 1 and this experiment is very similar, albeit with a little less variation across runs in the current experiment. Models still struggle in contexts where `yes' answers are expected. The fact that they seem to do better on the OR probe than the AND probe in this experiment does not necessarily mean that \textit{or} is easier to learn than \textit{and}, since as we noted in section \ref{sec:eval}, unlike with the other two contrasting function word pairs, \textit{and} and \textit{or} have very different input distributions. \textit{Or} is always used as a logical conjunct connecting referents in count questions, while \textit{and} is used in a much wider variety of question types, connecting different types of phrases. Some of the difficulty with AND probe questions in `yes' contexts may simply be due to the distribution over input questions the models see for \textit{and} and how different these questions are from our out-of-distribution probe questions. Frequency is clearly not the only factor at play determining how and when models come to learn these words.

\begin{figure}[H]
    \centering
    \includegraphics[height=6cm]{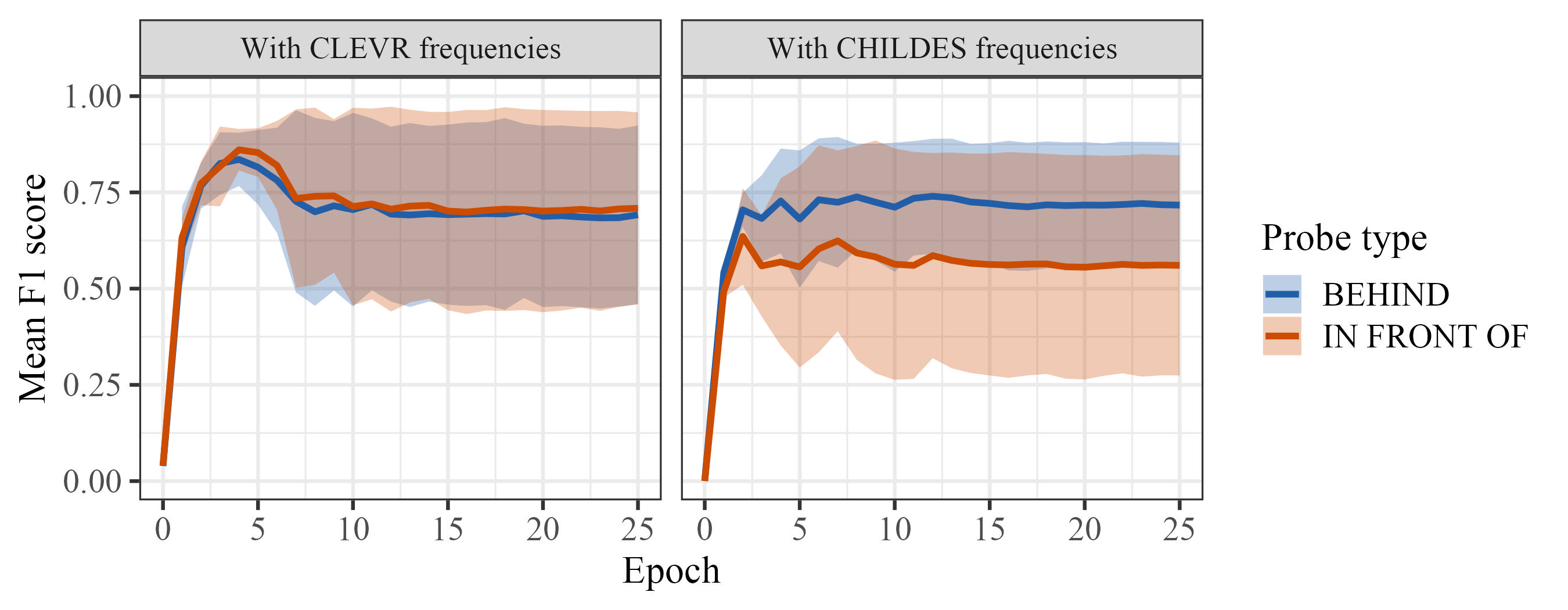}
    \caption[Experiment 3: F1 score on BEHIND - IN FRONT OF probes overall.]{Experiment 3: Mean F1 score on BEHIND - IN FRONT OF probes overall when trained on the original CLEVR dataset and the subsampled version with CHILDES-like frequencies. Shading represents standard deviation across 5 models.}
    \label{fig:behind-over-exp3}
\end{figure}

\paragraph{BEHIND - IN FRONT OF} These words are also not evenly distributed in children's input in CHILDES and consequently in our subsampled dataset. Both the number of instances of \textit{behind} and \textit{in front of} had to be reduced to create the training data used in this experiment, but we had to decrease the number of \textit{in front of} instances significantly more to reproduce their relative frequencies from CHILDES. As we can see in Figure \ref{fig:behind-over-exp3}, these changes had an effect on the overall performance of models on the IN FRONT OF probe which now finds itself on average around chance with much more variation across runs. The performance on the BEHIND probe is about the same in both conditions.

\begin{figure}[H]
    \centering
    \includegraphics[height=6cm]{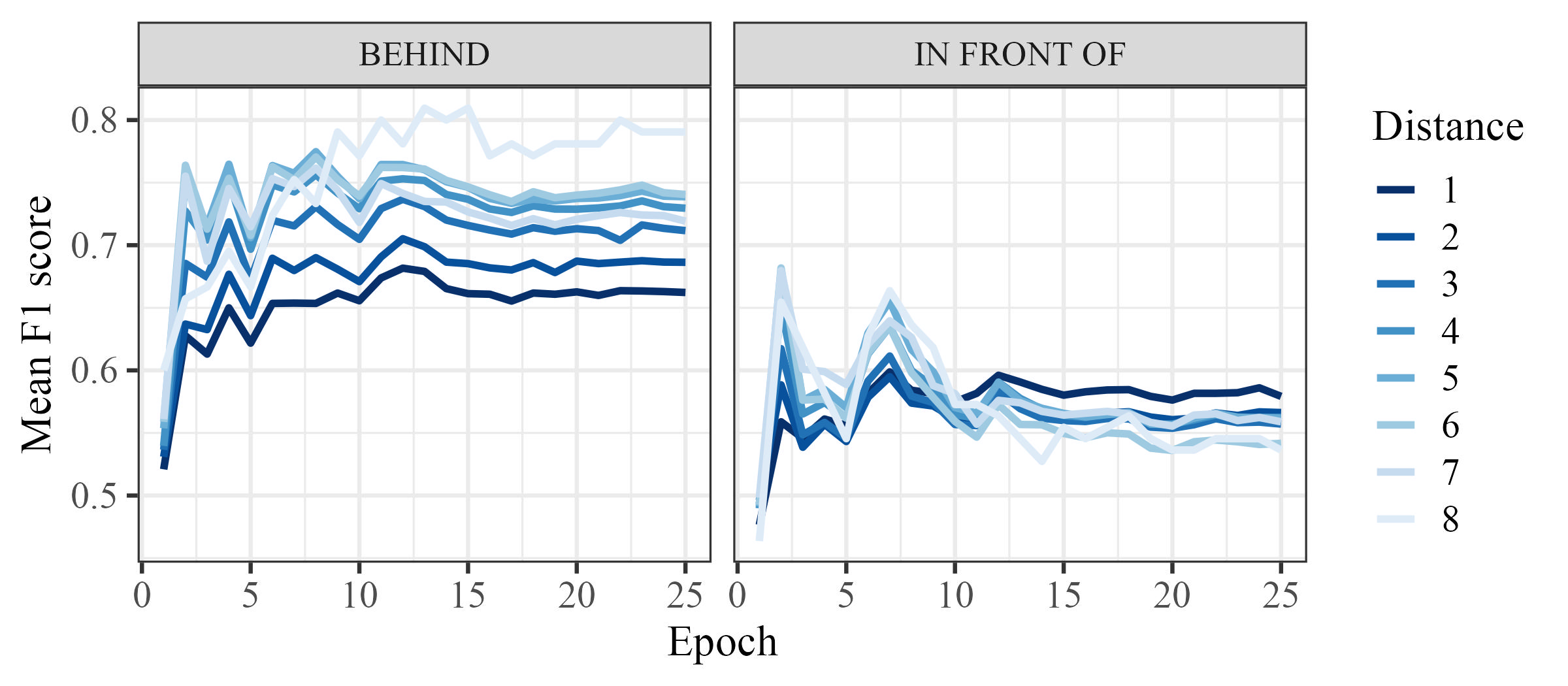}
    \caption[Experiment 3: F1 score on BEHIND - IN FRONT OF probes by distance.]{Experiment 3: Mean F1 score on BEHIND - IN FRONT OF probes as a function of the Euclidean distance between referents when trained of subsampled CLEVR.}
    \label{fig:behind-bydist-exp3}
\end{figure}

The most interesting results can be seen in Figure \ref{fig:behind-bydist-exp3} where we have plotted model performance on probe questions as a function of the Euclidean distance between the two referents in probe questions. Again, the results from experiment 1 for the BEHIND probe are reproduced, showing a clear gradience in interpretations of \textit{behind} as a function of distance. However, in the case of IN FRONT OF, the gradience has completely disappeared.

All of these results suggest that when models are trained on a CLEVR training dataset that reproduces the relative frequencies of \textit{behind} and \textit{in front of} seen in children's input, they learn the most frequent word of the pair, \textit{behind}, but struggle to learn the meaning of the less frequent opposing word, \textit{in front of}. This pattern differs from that of \textit{and} and \textit{or}, since for  \textit{behind} and \textit{in front of}, frequency does seem to be the most important factor in determining their relative learning order and difficulty.

\paragraph{MORE - FEWER} These words are an interesting case to consider because \textit{fewer} is extremely rare in children's input while \textit{more} is quite common. There are few different senses of the word \textit{more}, the most common in children's input being its adverbial form as in `do you want more?', which is quite different from the comparative quantifier \textit{more} seen in CLEVR as in `more than'. Since we could not easily differentiate all the senses of \textit{more}, we decided to also include its counterpart \textit{less} in addition to \textit{fewer} when determining their relative frequencies. Nonetheless, \textit{more} was much more frequent than \textit{fewer} and \textit{less} combined (see Table \ref{tab:clevr-childes-freq}).

\begin{figure}[H]
    \centering
    \includegraphics[height=6cm]{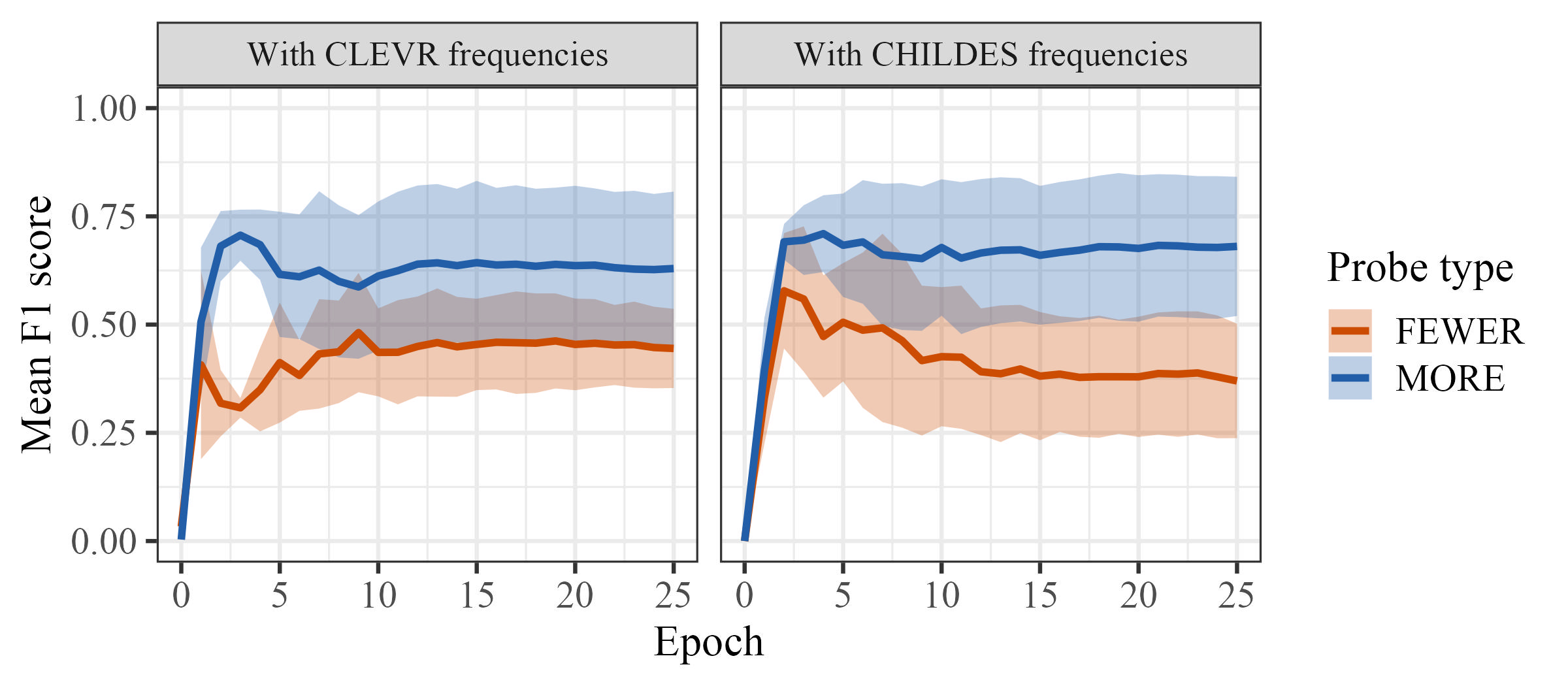}
    \caption[Experiment 3: F1 score on MORE - FEWER probes overall.]{Experiment 3: Mean F1 score on MORE - FEWER probes overall when trained on the original CLEVR dataset and the subsampled version with CHILDES-like frequencies. Shading represents standard deviation across 5 models.}
    \label{fig:more-over-exp3}
\end{figure}

Figure \ref{fig:more-over-exp3} compares the overall performance of models on both probes when trained on the original CLEVR dataset and the subsampled version with CHILDES-like frequencies. Performance on MORE is about the same, while on FEWER seems a little lower in the current experiment.

\begin{figure}[H]
    \centering
    \includegraphics[height=6cm]{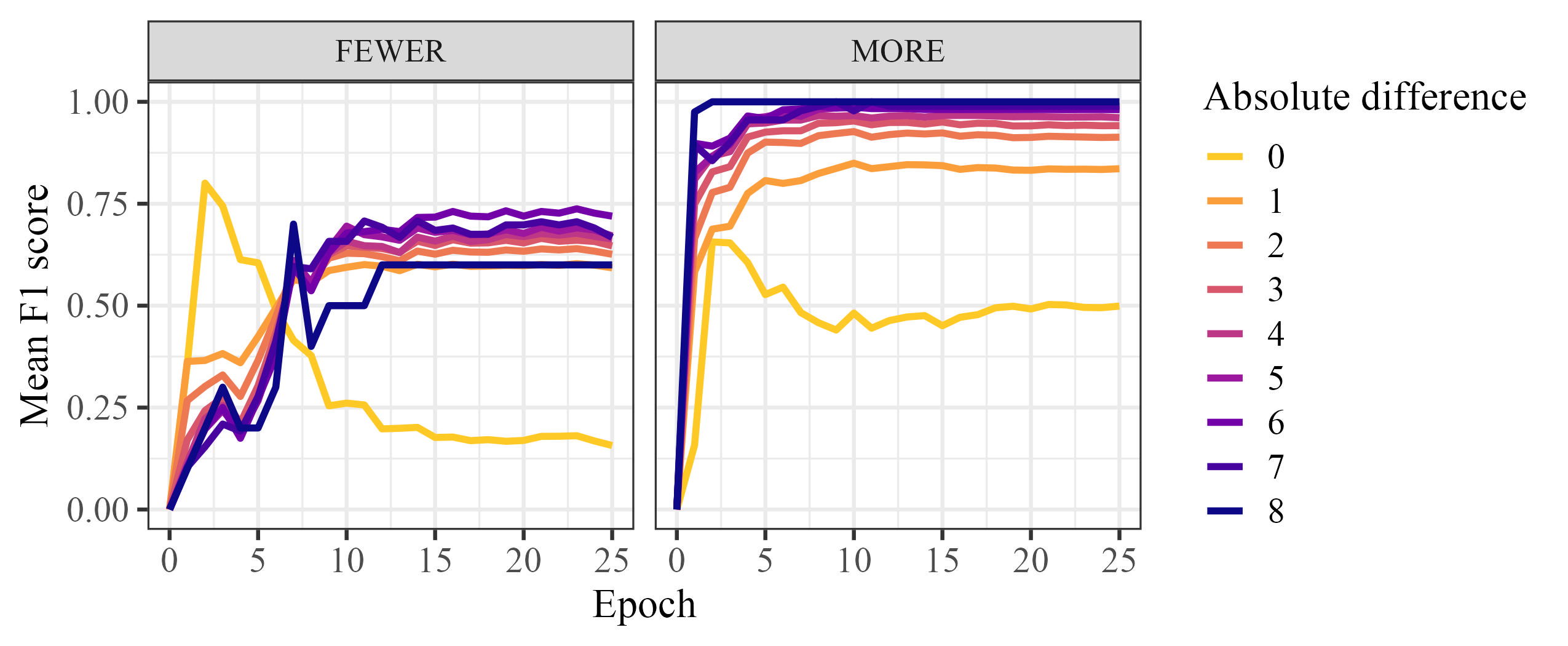}
    \caption[Experiment 3: F1 score on MORE - FEWER probes by absolute count difference.]{Experiment 3: Mean F1 score on MORE - FEWER probes by absolute difference in the number of objects in each referent class when trained of subsampled CLEVR.}
    \label{fig:more-bynum-exp3}
\end{figure}

Further probing with Figure \ref{fig:more-bynum-exp3} shows that errors are isolated specifically to contexts where $|X| = |Y|$ -- yet again. Surprisingly given the very small number of exemplars of \textit{fewer} seen during training -- only 105 cases -- models still seem to learn to use \textit{fewer} in unseen contexts as long as the absolute difference in number between referent classes is greater than zero. Additionally, unlike our results for \textit{in front of}, models still show some gradience in interpretation for \textit{fewer} as a function of number difference. Questions with \textit{fewer} are all answered with `yes' or `no', while questions with \textit{in front of} expect a much broader set of answers in the original CLEVR dataset (see Tables \ref{tab:clevr-freq} and \ref{tab:clevr-ans-freq}). This difference in input distribution might explain why models can still learn a reasonable representation for \textit{fewer} with so few examples.

\begin{figure}[H]
    \centering
    \includegraphics[height=6cm]{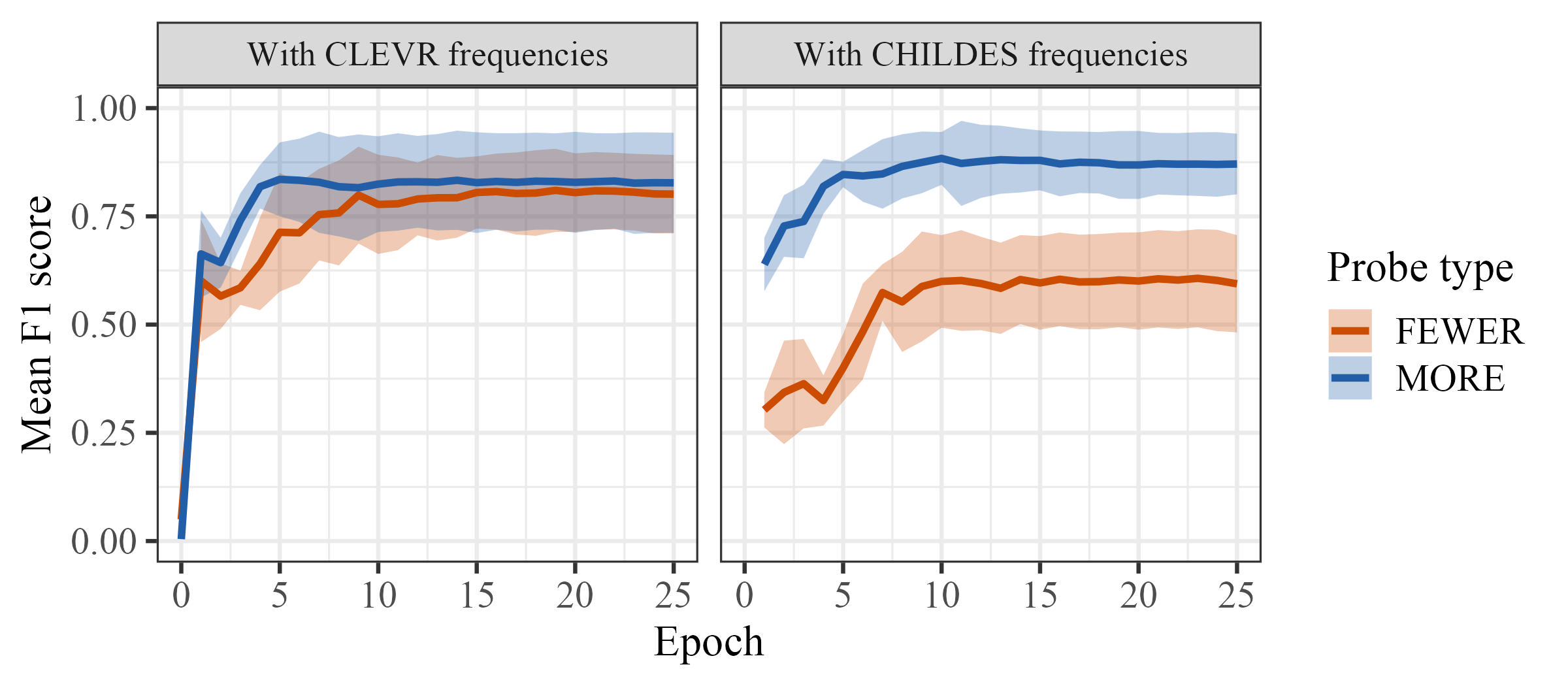}
    \caption[Experiment 3: F1 score on MORE - FEWER probes overall without even cases.]{Experiment 3: Mean F1 score on MORE - FEWER probes overall excluding context where $|X| = |Y|$, when trained on the original CLEVR dataset and the subsampled version with CHILDES-like frequencies. Shading represents standard deviation across 5 models.}
    \label{fig:more-overno0-exp3}
\end{figure}

By removing the probe questions where $|X| = |Y|$ and re-plotting models' F1 scores for all other cases in Figure \ref{fig:more-overno0-exp3}, we can clearly see that they learn to properly use both MORE and FEWER most of the time, though the performance on FEWER questions has definitely decreased in comparison to the results from experiment 1 when trained on the original CLEVR data.

Even with only a few exemplars of \textit{fewer}, models are able to learn reasonable meaning representations for this word, showing gradient interpretation as a function of the difference in number between compared classes. Models are not as accurate on the FEWER probe as their are on MORE questions. Once again, in contexts where $|X| = |Y|$, models struggle to answer both MORE and FEWER questions correctly. These results suggest that relative word frequency in the input also affects how models learn these function words.

\subsection{Interim conclusion}

With this experiment, we addressed our third research question: do models learn these function words in a similar order to children and are these ordering effects the results of their frequency or do they follow from other conceptual explanations? When trained on a corpus with similar relative frequencies to children's input, the MAC models had difficulty learning less frequent function words. For our logical reasoning targets, however, there are factors beyond frequency that influenced our models' ability to learn the meanings of \textit{and} and \textit{or}.

\section{General Discussion}

How children learn `hard' words like \textit{and/or}, \textit{behind/in front of}, and \textit{more/fewer} is still an open question. Proposals for their acquisition range along a spectrum between children having innate knowledge of the reasoning skills required to understand these words (a nativist perspective) to having to learn them from scratch using general learning mechanisms (a usage-based perspective). In this paper, we used a recurrent neural network model exposed to visually grounded language as a test-bed to evaluate the learnability of these function words, providing a proof-of-concept that such words can be learned from data.

First, we asked whether models were able to learn the meaning of these words using their non-symbolic general learning mechanisms. We found that they did learn to interpret function words along the way to succeeding in the visual question answering task they were trained on, the CLEVR dataset.  Models favored learning linguistic and visual representations that allowed for gradient interpretations for function words requiring spacial and numerical reasoning rather than threshold-based interpretations, showing that gradience in meaning may emerge from exposure to language in visually grounded contexts. Models also learnt to interpret logical connectives \textit{and} and \textit{or} without any prior knowledge of logical reasoning. Additionally, in answer to our second question, we found that models showed evidence of being sensitive to alternative possible expressions when inferring the meaning of these words, which lead to a rise in exclusive interpretations for \textit{or} in experiment 1. Finally, we wondered whether the relative difficulty of acquisition of words for children could be replicated in models and if it varied as a function of frequency rather than conceptual factors. We found that word learning difficulty was indeed dependent on word frequency in models' input, more frequently seen words generally being easier to learn in the case of spacial and numerical reasoning expressions. When exposed to these words at similar frequencies to children, models showed similar ordering effects for both \textit{behind/in front of} and \textit{more/fewer} word pairs. As for our logical reasoning targets, there seemed to be factors beyond frequency that influence our models' ability to learn them. One possible explanation for this difference may be that it is an artifact of the CLEVR dataset, which presented very different context distributions for \textit{and} and \textit{or} as opposed to other function word pairs.

We acknowledge that this work has its limitations. First, the CLEVR dataset is template-based and has a limited vocabulary. Its relative distribution of function words to content words like nouns and verbs is different from natural language, which may change the essence of what it means to be a function word or closed-class word. Additionally, the function words in question appear in a much wider variety of syntactic and semantic contexts in natural language. Though the dataset remains a good test bed for considering the acquisition of the reasoning skills necessary for interpreting these particular words in context, children acquiring these words may face challenges in naturalistic contexts that cannot be modeled with CLEVR data. Second, our probes do not allow us to determine where gradience in interpretations originates. We can only conclude that gradience arises from the integration of both visual and linguistic representations in the model. Third, our probes are a zero-shot evaluation looking at model generalization for a limited number of templates. To strengthen our conclusions, we would need to see the models response patterns extended to more templates. Still, this work exemplifies how we can probe the nuanced linguistic interpretations of visually grounded models for future studies.

It is possible to learn these complex and abstract reasoning skills and to map them to interpretations of function words without any prior knowledge. Our results offer proof of concept evidence that sophisticated statistical learning mechanism, when applied to visually grounded language, may be enough to explain the acquisition of these function words and related reasoning skills supporting more usage-based theories. Congruently, word learning difficulty was found to be mainly affected by frequency of exposure rather than conceptual factors.

Our work converges with other recent work suggesting that a variety of non-symbolic neural networks can learn logical operators from sufficiently rich data. For example, \citeauthor{geiger2023relational} (2023) showed that the logical operator ``same'' could be learned from data. Although our work here focused on a supervised learning regime, \citeauthor{geiger2023relational} showed learning successes across supervised and unsupervised contexts, supporting the idea that supervision does not necessarily play a key role in the emergence of symbolic structure. More broadly, the successes of large language models on large-scale reasoning tasks \cite{brown2020language,wei2022chain,kojima2022large} suggest that unsupervised learning may be sufficient for the emergence of functional representations supporting reasoning, though more work is needed to probe such models \cite{mahowald2023dissociating}.

The unprecedented success of neural network models offers an opportunity for cognitive science researchers to re-evaluate questions about the learnability of language \cite{lappin2021deep,warstadt2022what,piantadosi2023modern} and provides a new set of tools for comparisons between machine learning and child learning \cite{frank2019neural,portelance2023roles}.
We hope that our work here contributes to this broader enterprise.


\bibliographystyle{apacite}
\bibliography{vqa}

\newpage
\input{appendices_clean}

\end{document}

%% file: appendices_clean.tex



\appendix
\section{Templates from CLEVR dataset generator}\label{app:temp}
Here are the relevant templates from the CLEVR dataset generator for each function word. Words in brackets are options. The placeholders stand for: -S- is for shapes or other referent nouns; -M- are sets of adjectival modifiers (between 0 and 3 for size, color, material); -P- is for a property noun (``color", ``size", ``material", ``shape"); -R- is for a spacial relation (e.g. left of, right of, behind, above ). Note that the function words \textit{behind} and \textit{in front of} are considered relations in the CLEVR generative model and thus don't appear in the templates but can replace any -R- placeholder. They can therefore appear in some of these templates as well as many others.

\hfill

\noindent\textbf{\textit{And}}\\
\footnotesize
Are there an equal number of -M- -S-s and -M2- -S2-s?\\
Are there the same number of -M- -S-s and -M2- -S2-s?\\
Are there the same number of -M2- -S2-s [that are] -R- the -M- -S- and -M3- -S3-s?\\
Are there an equal number of -M2- -S2-s [that are] -R- the -M- -S- and -M3- -S3-s?\\
Are there an equal number of -M2- -S2-s [that are] -R- the -M- -S- and -M4- -S4-s [that are] -R2- the -M3- -S3-?\\
Are there the same number of -M2- -S2-s [that are] -R- the -M- -S- and -M4- -S4-s [that are] -R2- the -M3- -S3-?\\
Do the -M- -S- and the -M2- -S2- have the same -P-?\\
Are the -M- -S- and the -M2- -S2- made of the same material?\\
Do the -M2- -S2- [that is] -R- the -M- -S- and the -M3- -S3- have the same -P-?\\
Do the -M- -S- and the -M3- -S3- [that is] -R- the -M2- -S2- have the same -P-?\\
Do the -M2- -S2- [that is] -R- the -M- -S- and the -M4- -S4- [that is] -R2- the -M3- -S3- have the same -P-?\\
Are the -M2- -S2- [that is] -R- the -M- -S- and the -M3- -S3- made of the same material?\\
Are the -M- -S- and the -M3- -S3- [that is] -R- the -M2- -S2- made of the same material?\\
Are the -M2- -S2- [that is] -R- the -M- -S- and the -M4- -S4- [that is] -R2- the -M3- -S3- made of the same material?\\
How many -M3- -S3-s are [both] -R2- the -M2- -S2- and -R- the -M- -S-?\\
What number of -M3- -S3-s are [both] -R2- the -M2- -S2- and -R- the -M- -S-?\\
What is the -P- of the -M3- -S3- that is [both] -R2- the -M2- -S2- and -R- the -M- -S-?\\
What -P- is the -M3- -S3- that is [both] -R2- the -M2- -S2- and -R- the -M- -S-?\\
How big is the -M3- -S3- that is [both] -R2- the -M2- -S2- and -R- the -M- -S-?\\
There is a -M3- -S3- that is [both] -R2- the -M2- -S2- and -R- the -M- -S-; what is its -P-?\\
There is a -M3- -S3- that is [both] -R2- the -M2- -S2- and -R- the -M- -S-; what -P- is it?\\
There is a -M3- -S3- that is [both] -R2- the -M2- -S2- and -R- the -M- -S-; how big is it?\\
The -M3- -S3- that is [both] -R2- the -M2- -S2- and -R- the -M- -S- is what color?\\
What is the -M3- -S3- that is [both] -R2- the -M2- -S2- and -R- the -M- -S- made of?\\
The -M3- -S3- that is [both] -R2- the -M2- -S2- and -R- the -M- -S- is made of what material?\\
 There is a -M3- -S3- that is [both] -R2- the -M2- -S2- and -R- the -M- -S-; what [material] is it made of?\\
 The -M3- -S3- that is [both] -R2- the -M2- -S2- and -R- the -M- -S- has what shape?\\

\normalsize	
\noindent\textbf{\textit{Or}}\\
\footnotesize
How many [things/objects] are [either] -M- -S-s or -M2- -S2-s? \\
What number of [things/objects] are [either] -M- -S-s or -M2- -S2-s?\\
How many -P- [things/objects] are [either] -M- -S-s or -M2- -S2-s?\\
What number of -P- [things/objects] are [either] -M- -S-s or -M2- -S2-s?\\
How many [things/objects] are [either] -M2- -S2-s [that are] -R- the -M- -S- or -M4- -S4-s [that are] -R2- the -M3- -S3-?\\
What number of [things/objects] are [either] -M2- -S2-s [that are] -R- the -M- -S- or -M4- -S4-s [that are] -R2- the -M3- -S3-?\\
How many -S3-s are [either] -M- -S-s or -M2- -S2-s?\\
What number of -S3-s are [either] -M- -S-s or -M2- -S2-s?\\
How many [things/objects] are [either] -M2- -S2-s [that are] -R- the -M- -S- or -M3- -S3-s?\\
What number of [things/objects] are [either] -M2- -S2-s [that are] -R- the -M- -S- or -M3- -S3-s?\\
How many [things/objects] are [either] -M- -S-s or -M3- -S3-s [that are] -R- the -M2- -S2-?\\
What number of [things/objects] are [either] -M- -S-s or -M3- -S3-s [that are] -R- the -M2- -S2-?\\

\normalsize	
\noindent\textbf{\textit{More}}\\
\footnotesize
Are there more -M2- -S2-s [that are] -R- the -M- -S- than -M4- -S4-s [that are] -R2- the -M3- -S3-?\\
Are there more -Z- -C- -M- -S-s than -Z2- -C2- -M2- -S2-s?\\
Are there more -M2- -S2-s [that are] -R- the -M- -S- than -M3- -S3-s?\\

\normalsize	
\noindent\textbf{\textit{Fewer}}\\
\footnotesize
Are there fewer -M2- -S2-s [that are] -R- the -M- -S- than -M4- -S4-s [that are] -R2- the -M3- -S3-?\\
Are there fewer -M- -S-s than -M2- -S2-s?\\
Are there fewer -M2- -S2-s [that are] -R- the -M- -S- than -M3- -S3-s?\\

\normalsize	

\section{MAC model hyperparameters}\label{app:hyp3}
Here are the hyperparameters of the models we used for the experiments reported in the main paper. All models were trained on an Nvidia RTX 3080 GPU, taking about 15-20 hours including the probe evaluations every epoch. Like in the original paper \cite{hudson2018compositional}, models are trained using an Adam optimizer \cite{kingma2014adam}. Word vectors were initialized randomly using a standard uniform distribution based on a random seed. The models used a variational dropout of 0.15 across the network. We also used ELU as our non-linearity type.

\noindent\textit{number of epochs}: 25 \\
\textit{learning rate}: 0.0001, if performance on the validation set didn't improve from an epoch to the next, then the learning rate got reduced using a decay rate of 0.5.\\
\textit{batch size}: 64 \\
\textit{number of MAC cell layers}: 4 \\
\textit{hidden size}: 512 throughout network \\
\textit{embedding size}: 300 \\
\textit{image features output size}: 1024 \\
\textit{random seed}: [0:4]

\section{On models learning SAME} \label{app:same}

In initial versions of these experiments, we also included probes for the relational adjective \textit{same}. Though the word \textit{same} is part of the CLEVR vocabulary, the word \textit{different} is not. So instead of having a pair of opposing words for this reasoning skill, we considered how models learnt to interpret \textit{same} in contexts where objects did or did not share attributes. We tried two probe versions, one which tested the comparison of object sets and a second simpler one which required one-to-one object comparison. We chose not to include these results in the main paper as these probes proved to be either too hard or too easy for the models to answer, leading to overall uninformative results as to the learning dynamics of these models.

\subsection{The set comparison SAME probe results} 

This version of the SAME probe required models to consider sets of objects of varied sizes rather than to compare two specific referents. Questions were of the form `Are the $X$s the same \textit{property}', X being all possible referring expressions for objects within the CLEVR universe and \textit{property} being either \textit{color}, \textit{shape}, \textit{size} or \textit{material}. We matched possible properties to referents such that the value of these properties was not explicitly given by the referent terms (e.g. If the referents were \textit{brown sphere} and \textit{metal cube}, then the only possible value for property was \textit{size}, since the referent terms already mention the color, material, or shape of one of the objects). This first template presupposed that there were at least two $X$s. After finding all images for each question where this presuppositions was satisfied and then sampling 10 images if there were more than 10 available, this probe had a total of 1,170 image-question pairs.

In order for the answer to a question to be `yes' in the case of the first template all Xs in the image had to have the same value for the \textit{property} listed in the question, otherwise the answer was `no'. We used the `scene' metadata file associated with each image to verify whether this condition was met.

We ran this version with random seeds [0-2] before deciding to change to a different SAME question template, so the results presented here are based on 3 random runs rather than 5 like the rest of the experiments in the paper. We plot the mean F1 scores and standard deviation across runs. 

\begin{figure}[H]
    \centering
    \includegraphics[height=6cm]{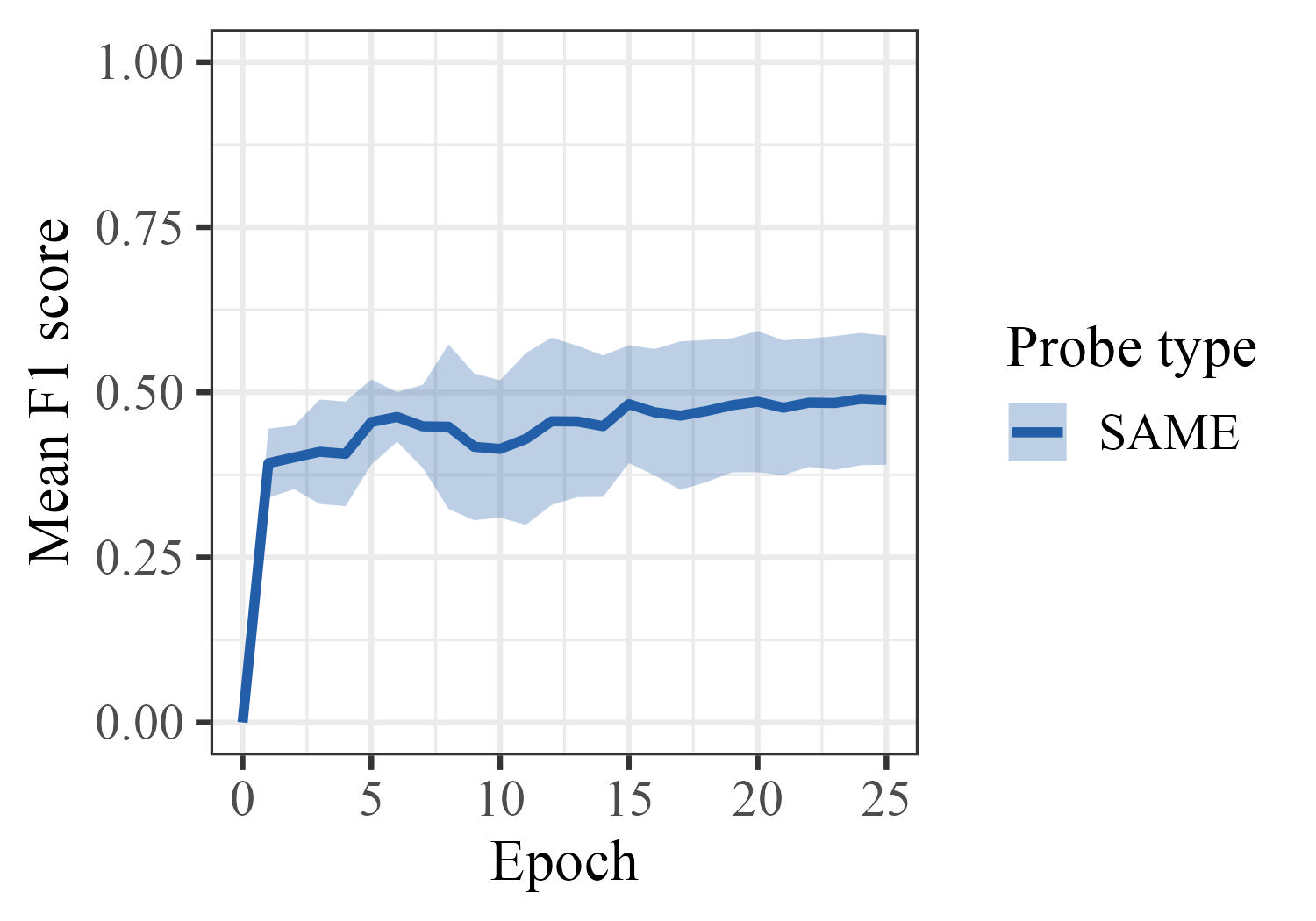}
    \caption{Mean F1 score on previous SAME probes overall. Shading represents standard deviation across 3 models.}
    \label{fig:oldsame-over}
\end{figure}

Figure \ref{fig:oldsame-over} shows the overall performance of the models on this version of the SAME probe. The models' performance is quite low.

\begin{figure}[H]
    \centering
    \includegraphics[height=6cm]{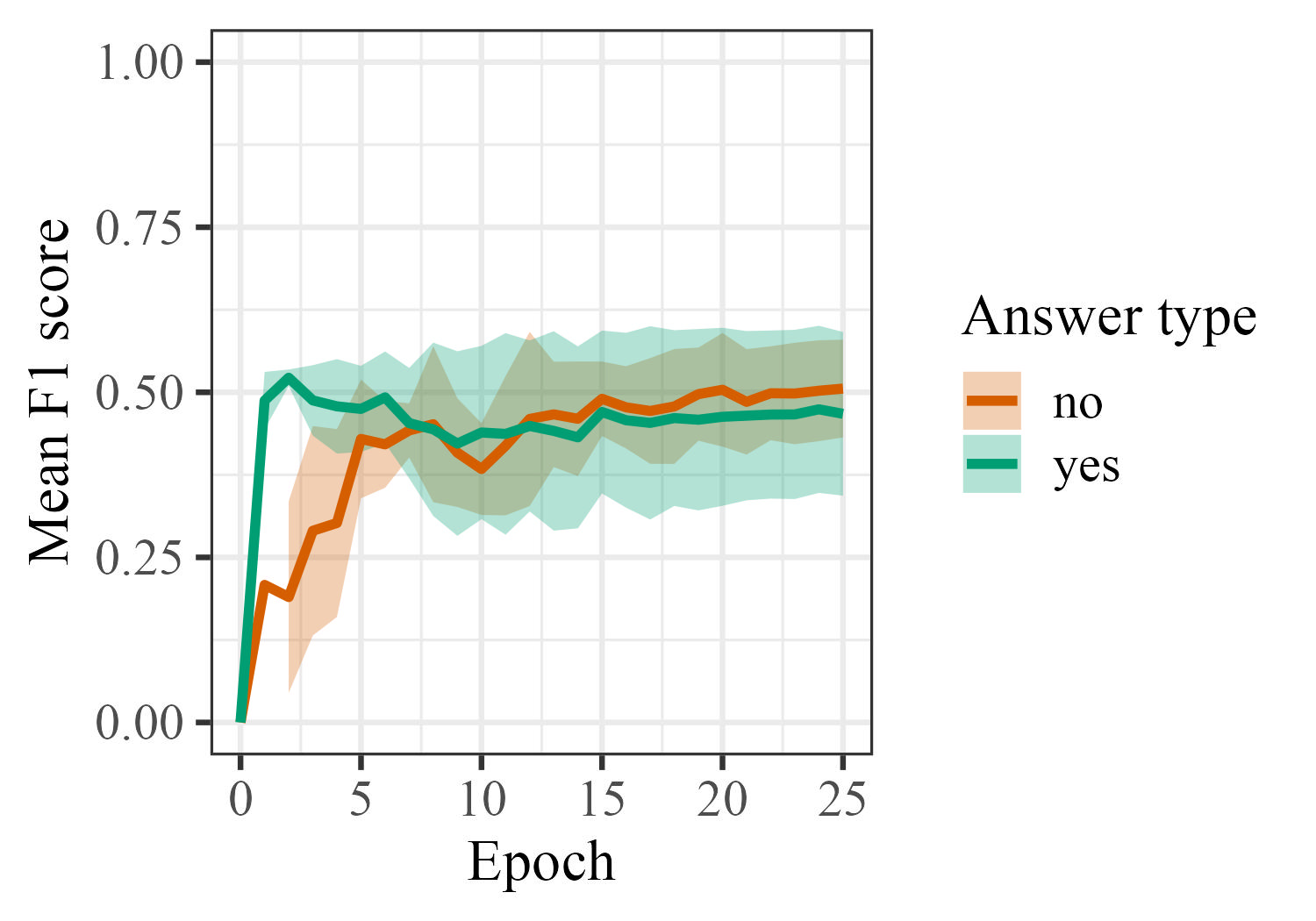}
    \caption[F1 score on previous SAME probes by answer type.]{Mean F1 score on previous SAME probes by answer type.}
    \label{fig:oldsame-ans}
\end{figure}

We can also look at results by answer type in Figure \ref{fig:oldsame-ans}. Again, it is clear that models really struggle to answer these questions as in either case they are only really considering two possible answers: "yes" or "no".

This initial version of the SAME probe may have been too different of a use of the word \textit{same} from those the model was exposed to during training, which could explain the models poor performance.

\subsection{The one-to-one comparison SAME probe results} 
\paragraph{SAME probes}

Given that models struggled with probe questions of the form `Are the $X$s the \textbf{same} \textit{property}?', we opted to try `Are the $X$ and the $Y$ the \textbf{same} \textit{property}?', where \textit{property} is one of the following: \textit{shape}, \textit{color}, \textit{material}, \textit{size}.  This second template presupposes that there are exactly one $X$ and one $Y$. After finding all images for each question where this presuppositions was satisfied and then sampling 10 images if there were more than 10 available, the second SAME probe had 38,580 question-image pairs. 

In order for the answer to a question to be `yes', $X$ and $Y$ had to share the same value for the given \textit{property} for a `yes' answer, otherwise the answer was  `no'.

We ran this version with random seeds [0-4], so the results presented here are based on 5 random runs like the rest of the experiments in the paper. We plot the mean F1 scores and standard deviation across runs. 

\begin{figure}[H]
    \centering
    \includegraphics[height=6cm]{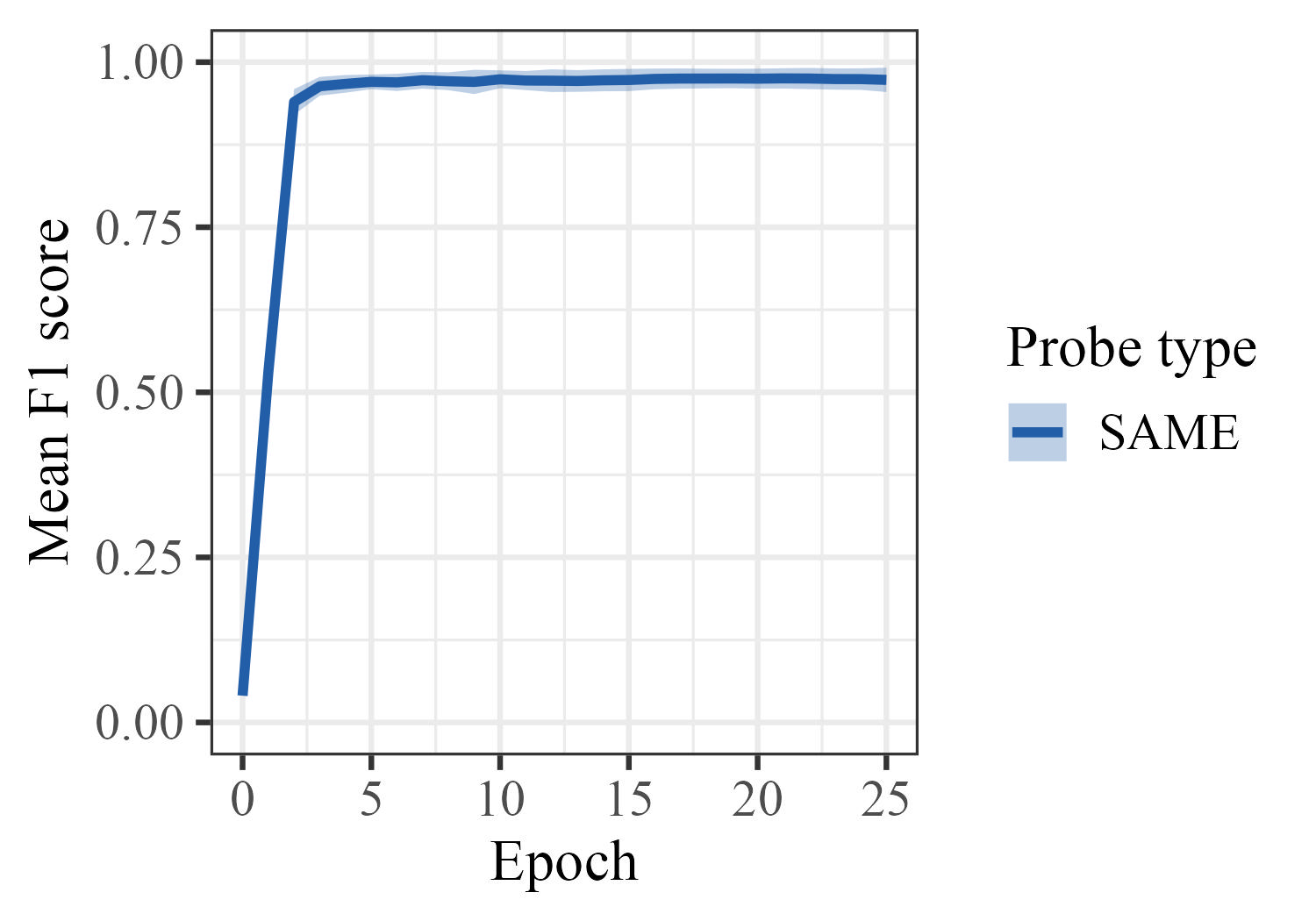}
    \caption[F1 score on SAME probe overall.]{ Mean F1 score on SAME probe overall. Shading represents standard deviation across 5 models.}
    \label{fig:same-over}
\end{figure}

Figure \ref{fig:same-over} shows the overall mean F1 scores of models on this SAME probe using the newer two referent comparison template instead. The models have almost perfect scores.

\begin{figure}[H]
    \centering
    \includegraphics[height=6cm]{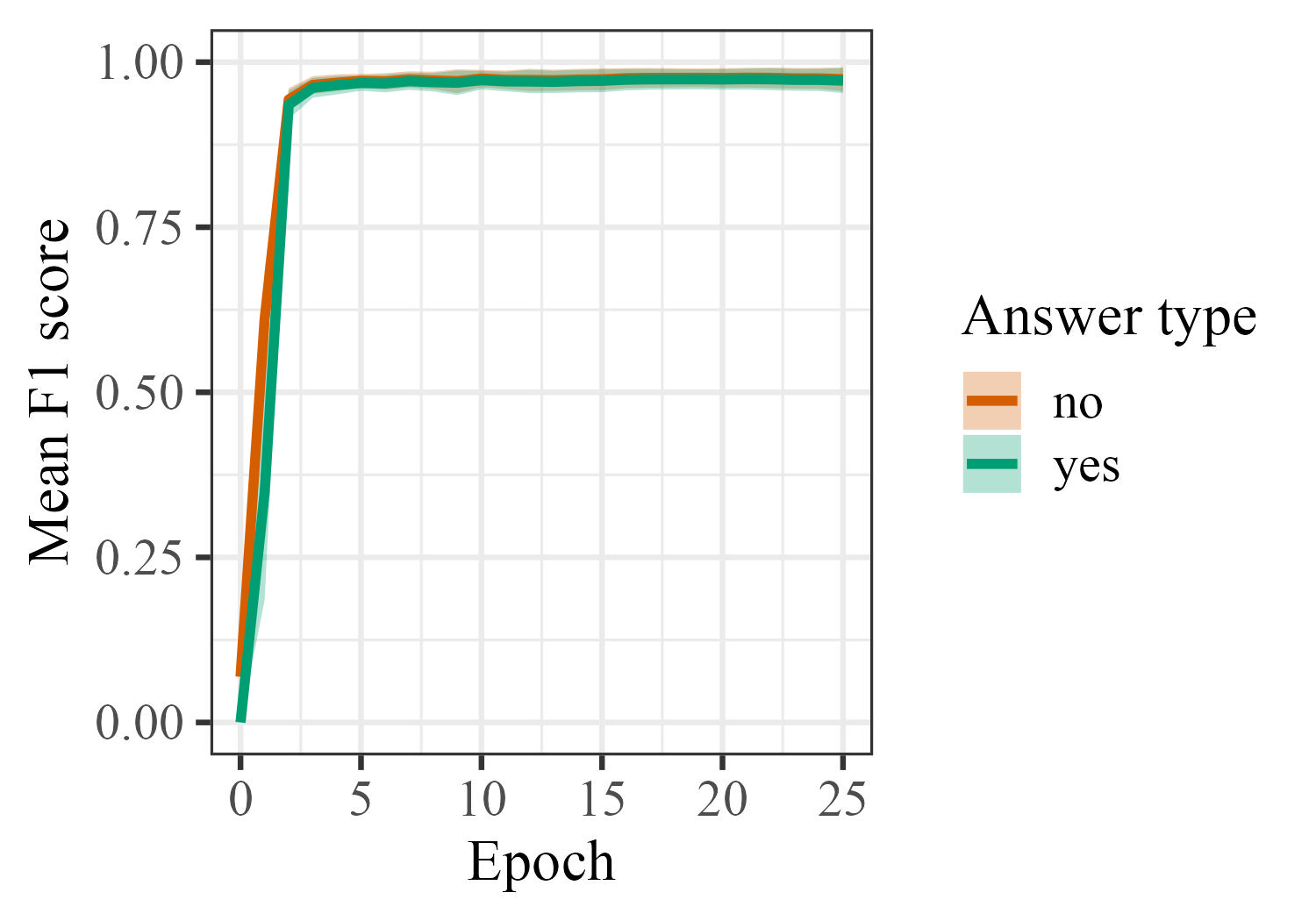}
    \caption{F1 score on SAME probe by answer type. Shading represents standard deviation across 5 models.}
    \label{fig:same-ans}
\end{figure}

When we considered models' performance as a function of the answer type expected, again we found that they score almost perfectly in either context, Figure \ref{fig:same-ans}.

Thus, though the models may struggle to generalize using \textit{same} to reason about items in a set, they learn to quickly use it to compare two unique referents. Additionally, they show no difference between correctly answering questions in contexts where these two referents share the \textit{same property} versus contexts where they do not. 

